%% file: neurips_2024.tex
\documentclass[pdftex]{article}

\PassOptionsToPackage{numbers, compress}{natbib}
\usepackage[final]{neurips_2024}

\usepackage{adjustbox}
\usepackage{multirow}
\usepackage{amsmath, amssymb, amscd, amsthm,mathrsfs, amsfonts}
\usepackage[utf8]{inputenc} % allow utf-8 input
\usepackage[T1]{fontenc}    % use 8-bit T1 fonts
\usepackage{hyperref}       % hyperlinks
\usepackage{url}            % simple URL typesetting
\usepackage{booktabs}       % professional-quality tables
\usepackage{amsfonts}       % blackboard math symbols
\usepackage{nicefrac}       % compact symbols for 1/2, etc.
\usepackage{microtype}      % microtypography
\usepackage[table,dvipsnames]{xcolor}  % for textcolor
\usepackage{graphicx}
\usepackage{graphics}
\usepackage[title]{appendix}
\usepackage{float}
\usepackage{makecell, cellspace, caption}
\usepackage{subcaption} % Required for creating subfigures
\usepackage{tikz}

\usepackage{xcolor}
\usepackage{tcolorbox}

\usepackage[dvipsnames]{xcolor}

\usepackage{makecell} % Include this line to avoid undefined control sequence error
\usepackage{multirow}
\usepackage{graphicx} % For resizebox
\usepackage{float}    % For [H] specifier
\usepackage{booktabs} % Optional: for better-looking tables

\definecolor{mygreen}{HTML}{80ac80}

\newcounter{oldtocdepth}

\newcommand{\hidefromtoc}{%
  \setcounter{oldtocdepth}{\value{tocdepth}}%
  \addtocontents{toc}{\protect\setcounter{tocdepth}{-10}}%
}

\newcommand{\unhidefromtoc}{%
  \addtocontents{toc}{\protect\setcounter{tocdepth}{\value{oldtocdepth}}}%
}

\definecolor{bluecolor}{HTML}{1f77b4}

\definecolor{redcolor}{HTML}{d62728}

\definecolor{color320}{HTML}{4878d0}
\definecolor{color384}{HTML}{ee854a}
\definecolor{color448}{HTML}{6acc64}

\newcommand{\tikzcircle}[2][redcolor,fill=redcolor]{\tikz[baseline=-0.5ex]\draw[#1,radius=#2] (0,0) circle ;}%

\title{Vision Transformer Neural Architecture Search for Out-of-Distribution Generalization: \\ ~Benchmark and Insights}

\author{%
  Sy-Tuyen Ho\thanks{Equal Contribution \hspace{5 mm} $^\dag$Corresponding Author} \hspace{5 mm} Tuan Van Vo$^*$ \hspace{5 mm} Somayeh Ebrahimkhani $^*$ \hspace{5 mm} Ngai-Man Cheung$^\dag$ \\
Singapore University of Technology and Design (SUTD) \\
\texttt{sytuyen\_ho@mymail.sutd.edu.sg},\\ \texttt{\{vovan\_tuan,somayeh\_ebrahimkhani,ngaiman\_cheung\}@sutd.edu.sg}
}

\begin{document}

\maketitle

\hidefromtoc

\input{sections/abstract}

\input{sections/introduction}

\input{sections/related_work}

\input{sections/benchmark}

\input{sections/analysis}

\input{sections/conclusion}

{\small
\bibliographystyle{unsrt}
\bibliography{egbib}
}

% \appendix

\input{sections/appendix}

% \newpage

% \input{sections/checklist}

% Optionally include supplemental material (complete proofs, additional experiments and plots) in appendix.
% All such materials \textbf{SHOULD be included in the main submission.}

%%%%%%%%%%%%%%%%%%%%%%%%%%%%%%%%%%%%%%%%%%%%%%%%%%%%%%%%%%%%

\newpage

\end{document}

%% file: sections/abstract.tex
\begin{abstract}
While Vision Transformer (ViT) have achieved success across various machine learning tasks, deploying them in real-world scenarios faces a critical challenge: generalizing under Out-of-Distribution (OoD) shifts. A crucial research gap remains in understanding how to design ViT architectures – both manually and automatically – to excel in OoD generalization. \textbf{To address this gap}, we introduce OoD-ViT-NAS, the first systematic benchmark for ViT Neural Architecture Search (NAS) focused on OoD generalization. This comprehensive benchmark includes $3,000$ ViT architectures of varying model computational budgets evaluated on $8$ common large-scale OoD datasets. With this comprehensive benchmark at hand, we analyze the factors that contribute to the OoD generalization of ViT architecture. Our analysis uncovers several key insights. Firstly, we show that ViT architecture designs have a considerable impact on OoD generalization. Secondly, we observe that In-Distribution (ID) accuracy might not be a very good indicator of OoD accuracy. This underscores the risk that ViT architectures optimized for ID accuracy might not perform well under OoD shifts. Thirdly, we conduct the first study to explore NAS for ViT's OoD robustness. Specifically, we study $9$ Training-free NAS for their OoD generalization performance on our benchmark. We observe that existing Training-free NAS are largely ineffective in predicting OoD accuracy despite their effectiveness at predicting ID accuracy. Moreover, simple proxies like \#Param or \#Flop surprisingly outperform more complex Training-free NAS in predicting ViTs OoD accuracy. Finally, we study how ViT architectural attributes impact OoD generalization. We discover that increasing embedding dimensions of a ViT architecture generally can improve the OoD generalization. We show that ViT architectures in our benchmark exhibit a wide range of OoD accuracy, with up to $11.85\%$ for some OoD shift, prompting the importance to study ViT architecture design for OoD. We firmly believe that our OoD-ViT-NAS benchmark and our analysis can catalyze and streamline important research on understanding how ViT architecture designs influence OoD generalization. {\bf Our OoD-NAS-ViT benchmark and code are available at \hyperlink{https://hosytuyen.github.io/projects/OoD-ViT-NAS}{https://hosytuyen.github.io/projects/OoD-ViT-NAS}}

\end{abstract}

%% file: sections/introduction.tex
\section{Introduction}

\begin{figure}[ht]
	\centering
	% Use the relevant command to insert your figure file.
	% For example, with the graphicx package use
	\includegraphics[width=.95\linewidth]{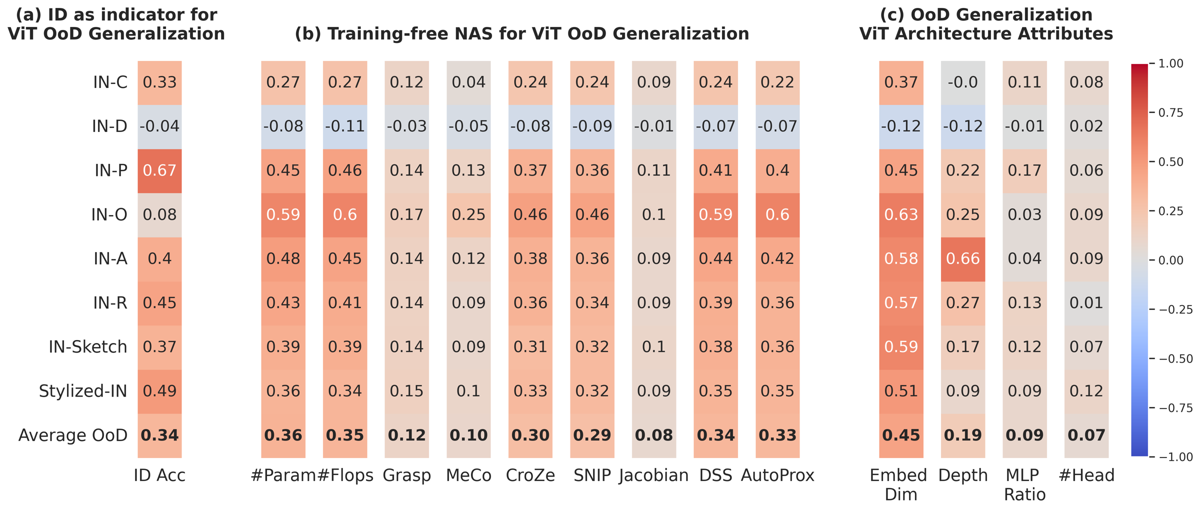}
	\vspace{-0.2cm}
	\caption{
   \textbf{We propose, OoD-ViT-NAS, the first comprehensive benchmark for NAS on OoD generalization of ViT architectures.} Then, we comprehensively investigate OoD generalization for ViT. The detailed of $8$ OoD datasets in our investigation can be found in Tab.~\ref{tab:benchmark_summary}.
   %Man
   In this figure, we show the 
   Kendall $\tau$ ranking correlation
   between OoD accuracy of different datasets on the left and different quantities at the bottom. 
   % The correlation is computed by Kendall $\tau$ ranking correlation. 
   Our analysis uncovers several key insights. \textbf{(a) ID as an indicator for ViT OoD Generalization (Sec.~\ref{Sec:IDvsOoDAcc})} We show that the correlation between ID accuracy and OoD accuracy is not very high. This suggests that current architectural insights based on ID accuracy might not translate well to OoD generalization. \textbf{(b) Training-free NAS for ViT OoD Generalization. (Sec.~\ref{Sec:Proxies})} We conduct the first study of NAS for ViT's OoD generalization, showing that their effectiveness significantly weakens in predicting OoD accuracy. \textbf{(c) OoD Generalization ViT Architectural Attributes. (Sec.~\ref{Sec:Robust_Arch})} Our first study on the impact of ViT architectural attributes on OoD generalization shows that the embedding dimension generally has the highest correlation with OoD accuracy among ViT architectural attributes. Additional results can be found in the Appx.}
   \vspace{-1.0cm}
	\label{fig:analysis-investigation}  
\end{figure}

Vision Transformers (ViT) \cite{dosovitskiy2020image} have recently achieved impressive results and become a major area of research in computer vision, with significant efforts towards understanding how ViT works. These efforts have led to the proposal of both manually designed architectures  \cite{dosovitskiy2020image,tu2022maxvit,liu2021swin,li2022exploring,zhang2022segvit} or automated-searched architectures \cite{chen2021autoformer,wang2023prenas,su2022vitas,tang2023elasticvit,zhou2022training,zhou2024training,gong2022nasvit} to advance ViT architectures. 

\textbf{Research Gap.} Existing research on ViT architectures focuses on maximizing In-Distribution (ID) accuracy, while studies on the impact of ViT architectures on Out-of-Distribution (OoD) generalization are limited. Initial works \cite{Bhojanapalli2021}, \cite{tang2021robustart}, and \cite{rahman2023out} evaluate sets of $3$, $10$ and $22$ human-designed ViT architectures under OoD settings, respectively, and provide coarse insights into which models exhibit better OoD generalization. However, with very limited ViT architectures studied in previous works, the influence of ViT structural attributes (e.g., embedding dimension, number of heads, MLP ratio, number of layers) on OoD generalization remains unclear. Besides, in the context of ViT Neural Architecture Search (NAS), while there are various ViT NAS for ID accuracy \cite{chen2021autoformer,wang2023prenas,chen2021searching,su2022vitas,zhou2022training,zhou2024training,wei2024auto,ding2021hr,liu2022uninet,yu2020bignas,tang2023elasticvit}, there is no study on ViT NAS for OoD generalization. 
% A more detailed discussion of related work can be found in the Appx.~\ref{Sec:Appx_Related_work}.

\textbf{In this paper,} we address existing research gaps by introducing OoD-ViT-NAS, the first comprehensive benchmark specifically designed for ViT's OoD generalization. Building NAS benchmarks is notoriously time-consuming and expensive due to the need to train and evaluate every candidate architecture. This challenge is particularly acute for ViT, known for its high computational demands and memory usage \cite{jiang2021all,chen2021crossvit,chen2021pre}. To overcome this bottleneck, we propose leveraging One-Shot NAS, specifically AutoFormer \cite{chen2021autoformer}, a widely used ViT search space.%We propose to overcome this bottleneck by leveraging One-Shot NAS, specifically Autoformer \cite{chen2021autoformer}, a widely used ViT search space. 
% \textcolor{RubineRed}{After training the Autoformer supernets}\somayeh{we are not training the supernets, we are using the pre-trained weights.}, 
We sample a diverse set of sub-architectures (models) to populate our benchmark. 
Importantly, these subnets inherit the weights from the pre-trained supernets, and \textit{their performance has been shown to be comparable to, or even superior to, that of architectures trained alone} \cite{yu2020bignas,chen2021autoformer}
% \textcolor{RubineRed}{Importantly, these subnets inherit the weights from the pre-trained supernets, and \textbf{it has shown} their performance to be comparable or even superior to architectures trained alone \cite{yu2020bignas,chen2021autoformer}.}\somayeh{Suggestion:  Notably, these subnets inherit the weights from the pre-trained supernets, which has been shown to result in performance comparable to or even superior to that of architectures trained from scratch.} 
This approach enables us to efficiently acquire a large pool of ViT architectures for OoD generalization analysis. Using OoD-ViT-NAS, we conduct extensive OoD generalization analysis and gain several key insights. Additionally, with our benchmark, our work is the first to explore (training-free) NAS for ViT's OoD generalization. Our contributions are summarized below:

\begin{itemize}
    \item We introduce OoD-ViT-NAS, the first comprehensive benchmark designed for NAS research on ViT's OOD generalization. This benchmark includes $3,000$ diverse ViT architectures sampled from the widely used ViT search space \cite{chen2021autoformer}. These architectures span a wide range of computational budgets. To thoroughly benchmark OoD generalization, these architectures are evaluated on the $8$ most common and state-of-the-art (SOTA) OoD datasets: ImageNet-C \cite{hendrycks2019benchmarking}, ImageNet-A \cite{hendrycks2021natural}, ImageNet-O \cite{hendrycks2021natural}, ImageNet-P \cite{hendrycks2019benchmarking}, ImageNet-D \cite{zhang2024imagenet}, ImageNet-R \cite{hendrycks2021many}, ImageNet-Sketch \cite{wang2019learning}, and Stylized ImageNet \cite{geirhos2018imagenet} (Sec.~\ref{Sec:Benchmark}) 

    \item Our analysis demonstrates the significant influence of ViT architectural designs on OoD accuracy. This observation encourages future research to focus more on ViT architecture research for OoD generalization (Sec.~\ref{sec:ood_accuracy_range})

    \item We show that high In-Distribution (ID) accuracy is not a very good indicator of OoD accuracy. This suggests that current architectural insights based on ID accuracy might not translate well to OoD generalization (Sec.~\ref{Sec:IDvsOoDAcc})

    \item We conduct the first study to explore NAS for ViT's OoD generalization. We study $9$ Training-free NAS for their OOD generalization performance on our benchmark. We observe that despite their prediction accuracy for ID, their effectiveness significantly weakens when predicting OoD accuracy. Furthermore, simple proxies such as the number of parameters (\#Param) or the number of floating point operations (\#Flop) surprisingly outperform more complex Training-free NAS in predicting ViTs OoD accuracy (Sec.~\ref{Sec:Proxies})

    \item We study the impact of ViT architecture design on OoD generalization and demonstrate that careful design of ViT architectures can significantly improve OoD generalization. Specifically, increasing the embedding dimensions of a ViT architecture generally can improve its OoD generalization. (Sec.~\ref{Sec:Robust_Arch}) We show that architectures with comparable ID accuracy (within an averaging range of $1.39\%$) exhibit a wider range of OoD accuracy, averaging $3.80\%$ and reaching highs of $11.85\%$, being comparable or even outperforming state-of-the-art (SOTA) training OoD generalization method, such as those based on domain invariant representation learning \cite{bai2024hypo,arjovsky2019invariant}. For example, under the same OoD setting, the SOTA method \cite{bai2024hypo} shows an improvement of $1.9\%$ OoD accuracy. 
\end{itemize}

%% file: sections/related_work.tex
\section{Related Work}

% \textbf{Vision Transformers (ViTs).} ViTs are a rapidly evolving deep learning architecture in the computer vision domain. Unlike Convolutional Neural Networks (CNNs), ViTs leverage attention mechanisms, originally developed for natural language processing \cite{vaswani2017attention}, to capture long-range dependencies within images. This architecture relies on two key components: Channel-wise MLP blocks and Attention blocks. Recently, ViT architectures \cite{dosovitskiy2020image,tu2022maxvit,liu2021swin,chen2021autoformer,wang2023prenas,li2022exploring,zhang2022segvit} have demonstrated impressive performance across various computer vision tasks.

\textbf{Out-of-Distribution (OoD) Generalization}. Addressing Out-of-distribution (OoD) generalization is a challenge, particularly in computer vision. Various approaches have been proposed to tackle this issue. A common strategy focuses on learning features that remain consistent across different domains, thereby promoting generalizability  \cite{arjovsky2019invariant,ghifary2015domain,mahajan2021domain,muandet2013domain,rame2022fishr,rojas2018invariant,shi2021gradient}. Other directions explore distributional robustness \cite{sagawa2019distributionally,zhou2020learning}, model ensembles \cite{chen2023explore,rame2023model}, test-time adaptation \cite{park2023test,chen2023improved}, data augmentation techniques \cite{nam2021reducing,chen2022self,kim2021selfreg,li2021simple,zhou2021domain,xu2021fourier,yan2020improve}, and meta learning \cite{du2020metanorm,shu2021open} for OoD generalization. From an architectural perspective, a few attempts investigate the impact of network architecture on OoD generalization. Early work \cite{sagawa2020investigation} shows that over-parameterized networks can hinder OoD performance due to overfitting. This raises an intriguing question: can sub-networks within such architectures achieve better OoD performance? Inspired by the Lottery Ticket Hypothesis (LTH) \cite{frankle2018lottery}, the Functional LTH has been explored and shown that over-parameterized networks harbor sub-networks with better OoD performance. Techniques like Modular Risk Minimization \cite{zhang2021can} and Debiased Contrastive Weight Pruning \cite{park2023training} aim to identify these winning tickets. Another direction \cite{bai2021ood,tang2021robustart} leverages Neural Architecture Search (NAS) to analyze the OoD robust architectures. However, these studies primarily focus on CNNs. While ViTs have achieved success in various visual recognition, investigations into their OoD generalization are limited. Initial works \cite{Bhojanapalli2021}, \cite{tang2021robustart}, and \cite{rahman2023out} evaluate sets of $3$, $10$, and $22$ human-designed ViT architectures respectively, under OoD settings. Their results provide coarse insights into which models exhibit better OoD performance. \textit{However, the influence of ViT architectural attributes on OoD robustness remains unclear.}

\textbf{Neural Architecture Search (NAS)}. NAS is a promising approach that has achieved remarkable success in automatically searching efficient and effective architectures for ID performance \cite{wang2021alphanet,yu2020bignas,wang2021attentivenas,jiang2024meco}. Recently, NAS has been explored in the context of adversarial robustness for CNNs as well \cite{jung2023neural,ha2024generalizable,wurobust}. With the rise of Vision Transformers (ViTs), several NAS approaches have been applied to improve ViT architectures, including Autoformer \cite{chen2021autoformer}, S3 \cite{chen2021searching}, ViTAS 
\cite{su2022vitas}, ElasticViT \cite{tang2023elasticvit}, DSS \cite{zhou2022training}, Auto-Prox \cite{wei2024auto} and GLiT \cite{chen2021glit}. Additionally, hybrid CNN-ViT architectures like HR-NAS \cite{ding2021hr}, UniNet \cite{liu2022uninet}, and NASViT \cite{gong2022nasvit} have also been explored. These efforts have shown promising results in terms of ID accuracy. \textit{However, there has not been any work on NAS for ViT architectures specifically for OoD generalization.}

%% file: sections/benchmark.tex
\begin{figure}[t]
	\centering
	\includegraphics[width=.84\linewidth]
 {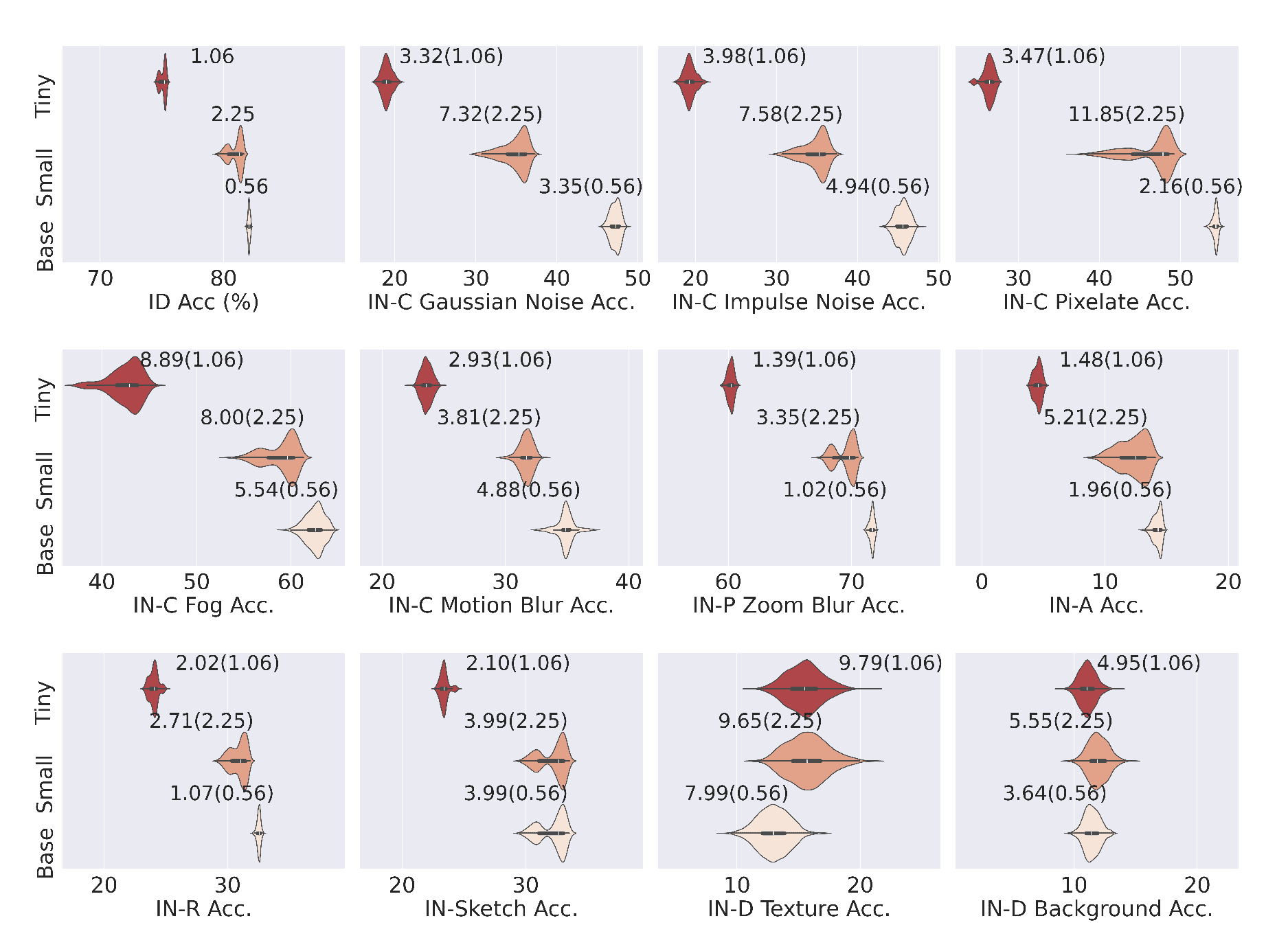}
 \vspace{-0.2cm}
    \caption{\textbf{Our analysis of the OoD accuracy range highlights the significant influence of ViT architectural designs on OoD accuracy. (Sec.~\ref{sec:ood_accuracy_range})} The numbers within each violin plot for each sub-figure (e.g., IN-D $9.79$ ($1.06$), $9.65$ ($2.25$), and $7.99$ ($0.56$)) denote the corresponding OoD (ID) accuracy range of architectures sampled from Autoformer-Tiny/Small/Base search space, respectively. See Appx. \ref{Sec:Appx_OoDrange} for additional plots and results on other OoD shifts. For a fair comparison, we fix the same range for the x-axis across all sub-figures. We include the ID accuracy range in the top-left sub-figure for reference. On average, the OoD accuracy across all shifts is $3.8\%$/$4.86\%$/$2.74\%$ for the search spaces in our OoD-ViT-NAS benchmark. This range is comparable to and even surpasses the current SOTA method based on domain-invariant representation learning \cite{bai2024hypo}, which achieved a $1.9\%$ improvement in OoD accuracy under similar settings.}
\vspace{-1.0cm}
	 
	\label{fig:OoDshifts_Visualization}
\end{figure}

\section{OoD-ViT-NAS: NAS Benchmark for ViT's OoD Generalization} \label{Sec:Benchmark}

In this section, we describe the construction of our OoD-ViT-NAS benchmark with details on the search spaces, datasets, evaluation metrics, and protocol. Our comprehensive benchmark includes $3,000$ ViT architectures of varying sizes evaluated on $8$ widely used large-scale, high-resolution, and SOTA OoD datasets. Our OoD-ViT-NAS benchmark is summarized in the Tab.~\ref{tab:benchmark_summary}

\textbf{Search Space.} We construct our benchmark based on Autoformer \cite{chen2021autoformer} search space. This search space is currently a widely used search space in the ViT NAS community for ID data \cite{wang2023prenas,zhang2023shiftnas,jiang2024meco,zhou2022training,zhou2024training,wei2024auto,wang2023mdl,li2023efficient}. Autoformer search space is a large vision transformer search space including five architectural attributes that define the building block. \textit{Embedding Dimension:} This determines the input feature representation size and is typically consistent across layers in ViT architectures. \textit{Q-K-V Dimension:} This specifies the size of the query, key, and value vectors used in the attention mechanism. \textit{Number of Heads:} This defines the number of parallel attention computations performed within a single attention block. \textit{MLP Ratio:} This controls the dimensionality of the feed-forward network within each transformer block. Unlike embedding dimension, in Autoformer search space, Q-K-V Dimension, Number of Heads, and MLP Ratio can be varied across layers. \textit{Network Depth:} This refers to the total number of transformer layers stacked in the architecture. It is important to note that Autoformer maintains a fixed ratio between the Q-K-V dimension and the number of heads in each block. This ensures that the scaling factor in the attention calculation remains constant. This helps stabilize the gradients of different heads during the training \cite{chen2021autoformer}. We strictly follow Autoformer search space. The details can be found in the Appx.~\ref{Sec:Appx_Benchmark_Searchspace}.

% \begin{wraptable}{r}{5cm}
\begin{table}[ht!]
\vspace{-0.7cm}
\caption{\textbf{An overview of comprehensive setups to construct our OoD-ViT-NAS benchmark.} We utilize the widely used ViT NAS search space, Autoformer \cite{chen2021autoformer}, which includes three different search spaces Autoformer-Tiny/Small/Base to cover a broad range of model sizes. We randomly sample \textbf{$3,000$ architectures} from these search spaces to populate our benchmark. To ensure comprehensiveness, we evaluate these architectures across 8 of the most common SOTA OoD datasets. Following prior OoD generalization works \cite{bai2021ood,bai2024hypo,arjovsky2019invariant,rahman2023out}, we employ three metrics for our benchmark: ID Accuracy, OoD Accuracy, and Area Under the Precision-Recall Curve (AUPR).}
\begin{adjustbox}{width=0.85\textwidth, center}
\renewcommand{\arraystretch}{1.4}
\begin{tabular}{llcccll}
\hline
\textbf{Search Space}                                                                                                 & \textbf{Dataset}  & \textbf{\#Classes} & \textbf{\#Images} & \textbf{\#OoD Shifts} & \textbf{OoD Shift Type}      & \textbf{Metrics}                                                           \\ \hline
\multirow{9}{*}{\begin{tabular}[c]{@{}l@{}}Autoformer-Tiny \cite{chen2021autoformer}/\\ \\ Autoformer-Small \cite{chen2021autoformer}/\\ \\ Autoformer-Base \cite{chen2021autoformer} \end{tabular}} & ImageNet-1K (IN-1K) \cite{fei2009imagenet}      & 1K                 & 50K               & -                     & -                            & -                                                                          \\ \cline{2-7} 
                                                                                                                      & ImageNet-C (IN-C) \cite{hendrycks2019benchmarking}        & 1K                 & 3.75M             & 15                    &\multirow{4}{*}{Algorithmic} & \multirow{7}{*}{\begin{tabular}[c]{@{}l@{}}ID Acc/\\ OoD Acc\end{tabular}} \\ \cline{2-5}
                                                                                                                      & ImageNet-P (IN-P) \cite{hendrycks2019benchmarking}        & 1K                 & 1M                 & 9                     &                              &                                                                            \\ \cline{2-5}
                                                                                                                      & ImageNet-D (IN-D) \cite{zhang2024imagenet}       & 113                & 4.8K              & 3                     &                              &                                                                            \\ \cline{2-5}
                                                                                                                      & Stylized ImageNet (Stylized IN) \cite{geirhos2018imagenet} & 1K                 & 50K               & 1                     &                              &                                                                            \\ \cline{2-6}
                                                                                                                      & ImageNet-R (IN-R) \cite{hendrycks2021many}       & 200                & 30K               & 1                     & \multirow{4}{*}{Natural}     &                                                                            \\ \cline{2-5}
                                                                                                                      & ImageNet-Sketch (IN-Sketch) \cite{wang2019learning}  & 1K                 & 50K               & 1                     &                              &                                                                            \\ \cline{2-5}
                                                                                                                      & ImageNet-A (IN-A)  \cite{hendrycks2021natural}      & 200                & 7.5K                 & 1                     &                              &                                                                            \\ \cline{2-5} \cline{7-7} 
                                                                                                                      & ImageNet-O (IN-O) \cite{hendrycks2021natural}       & 200                & 2K                 & 1                     &                              & \begin{tabular}[c]{@{}l@{}}ID Acc/\\ AUPR\end{tabular}                     \\ \hline
\end{tabular}
\end{adjustbox}
\vspace{-0.3cm}
\label{tab:benchmark_summary}
\end{table}

\textbf{Dataset.} Our benchmark consists of the evaluation on large-scale, high-resolution, and most SOTA OoD datasets, including ImageNet-1k \cite{fei2009imagenet}, ImageNet-C \cite{hendrycks2019benchmarking}, ImageNet-P \cite{hendrycks2019benchmarking}, ImageNet-A \cite{hendrycks2021natural}, ImageNet-O \cite{hendrycks2021natural}, ImageNet-R \cite{hendrycks2021many}, ImageNet-Sketch \cite{wang2019learning}, Stylized ImageNet \cite{geirhos2018imagenet}, and ImageNet-D \cite{zhang2024imagenet}. These datasets capture a comprehensive range of OoD shifts such as common corruptions (blur, noise, digital, weather), Stable-Diffusion-based OoD shifts, and natural OoD shifts. A detailed description of these datasets can be found in the Appx.~\ref{Sec:Appx_Benchmark_Dataset}.

\textbf{Metrics.} Following the previous OoD generalization methods \cite{bai2021ood,bai2024hypo,arjovsky2019invariant,rahman2023out}, we employ three metrics to construct our benchmark:

\begin{itemize}
    \item \textit{ID Classification Accuracy (ID Acc)}: This metric measures the model performance on In-Distribution (ID) data, typically the data it was trained on (e.g., ImageNet). A higher ID Acc indicates the model's ability to learn training data's distribution.

    \item \textit{OoD Classification Accuracy (OoD Acc)}: This metric measures the model performance on Out-of-Distribution (OoD) data, which could differ significantly from the training data. A higher OoD Acc indicates a better generalization of the model to handle the OoD shifts.

    \item For the specific case of ImageNet-O, \cite{hendrycks2019benchmarking}, we use the Area Under the Precision-Recall Curve (AUPR) metric. A higher AUPR indicates a better generalization of the model to handle the OoD detection.
\end{itemize}

\textbf{Protocol.} 
Neural Architecture Search (NAS) is notorious for its computationally expensive nature, requiring the training and evaluation of numerous candidate architectures. To address this challenge and efficiently obtain the large number of architectures needed for our benchmark (i.e., $3,000$), we make use of the One-Shot NAS approach \cite{wang2021alphanet,yu2020bignas,wang2021attentivenas,gong2022nasvit,chen2021autoformer,wang2023prenas}.

In One-shot NAS, a single supernet is first constructed. This supernet contains all possible architectures within the defined search space and is trained only once. Then, during evaluation, various architectures (i.e., subnets) can be efficiently extracted from the supernet. Importantly, these subnets inherit the weights from the pre-trained supernet, and their performance has been shown to be comparable or even superior to that of architectures trained alone \cite{yu2020bignas,chen2021autoformer}.

To support a wide range of model sizes, we leverage three supernets: Autoformer-Tiny/Small/Base, which were previously proposed for ID accuracy \cite{chen2021autoformer}. We randomly sample $1,000$ architectures from each supernet, resulting in a total of $3,000$ architectures in our OoD-ViT-NAS benchmark. Once obtained, these architectures are evaluated on 8 aforementioned OoD datasets.

% \begin{figure}[t]
% 	\centering
% 	%\includegraphics[width=14cm]{figures/violin_acc_12.png}
% 	\includegraphics[width=1.\linewidth]
%  %\includegraphics[width=14cm]
%  {figures/violin_acc_12.png}
%  \vspace{-0.5cm}
%     \caption{\textbf{Our analysis of the OoD accuracy range highlights the significant influence of ViT architectural designs on OoD accuracy. (Sec.~\ref{sec:ood_accuracy_range})} The numbers within each violin plot for each sub-figure (e.g., IN-D 9.79 (1.06), 9.65 (2.25), and 7.99 (0.56)) denote the corresponding OoD(ID) accuracy range of architectures sampled from Autoformer-Tiny/Small/Base search space, respectively. See Appx. \ref{Sec:Appx_OoDrange} for additional plots and results on other OoD shifts. For a fair comparison, we fix the same range for the x-axis across all sub-figures. We include the ID accuracy range in the top-left sub-figure for reference. On average, the OoD accuracy across all shifts is 3.8\%/4.86\%/2.74\% for the search spaces in our OoD-ViT-NAS benchmark. This range is comparable to and even surpasses the current SOTA method based on domain-invariant representation learning \cite{bai2024hypo}, which achieved a 1.9\% improvement in OoD accuracy under similar settings.}
% \vspace{-0.5cm}
	 
% 	\label{fig:OoDshifts_Visualization}
% \end{figure}

%% file: sections/analysis.tex
\section{Investigation on Out-of-Distribution Generalization of ViT} \label{Sec:Investigation}

In this section, we provide the first comprehensive investigation of how ViT architectures affect OoD generalization using our OoD-ViT-NAS benchmark. In Sec.~\ref{sec:ood_accuracy_range}, we first demonstrate that ViT architectures considerably impact OoD accuracy. In Sec.~\ref{Sec:IDvsOoDAcc}, while existing works \cite{tu2022maxvit,liu2021swin,chen2021autoformer,wang2023prenas,su2022vitas,tang2023elasticvit,gong2022nasvit,ding2021hr} have made significant strides in improving ViT's ID accuracy, their findings could not be applicable for ViT's OoD generalization due to the not very high correlation between ViT's ID and OoD accuracy. In Sec.~\ref{Sec:Proxies}, we conduct the first study to explore NAS for ViT's OoD generalization. Specifically, we study $9$ Training-free NAS based on their OoD generalization performance on our benchmark. Finally, in Sec.~\ref{Sec:Robust_Arch}, we analyze the influence of individual ViT architectural attributes (i.e., embedding dimension, number of heads, MLP ratio, number of layers) on OoD generalization. Additional results of these analysis can be found in Appx.~\ref{Sec:Appx_OoDrange},~\ref{Sec:Appx_IDvsOoD},~\ref{Sec:Appx_Pareto},~\ref{Sec:Appx_Proxies},~\ref{Sec:Appx_ViTAttribute}

% We summarize our investigation on OoD generalization of ViT in Fig.~\ref{fig:analysis-investigation} 

% To get the first overview of the relations between architectural design and their OoD generalization, we perform an extensive set of experiments comparing the performance of various transformer architectures. This allows us to identify the relationship between a model's performance on in-distribution (ID) data and its accuracy on OoD data under different distribution shifts. Second, we focus on key design elements within transformers. Through correlation analysis and ablation studies with extended settings, we analyze how these factors influence OoD robustness and generalization. Additionally, we evaluate the robustness of 3,000 ViT architectures against diverse OoD shifts using nine zero-cost proxies in our benchmark.

% \vspace{-1.0cm}
\subsection{ViT architecture designs have a considerable impact on OoD generalization} \label{sec:ood_accuracy_range}

\begin{figure}[t]
	\centering
	% Use the relevant command to insert your figure file.
	% For example, with the graphicx package use
	% \vspace{-0.4cm}
 \includegraphics[width=0.85\linewidth]{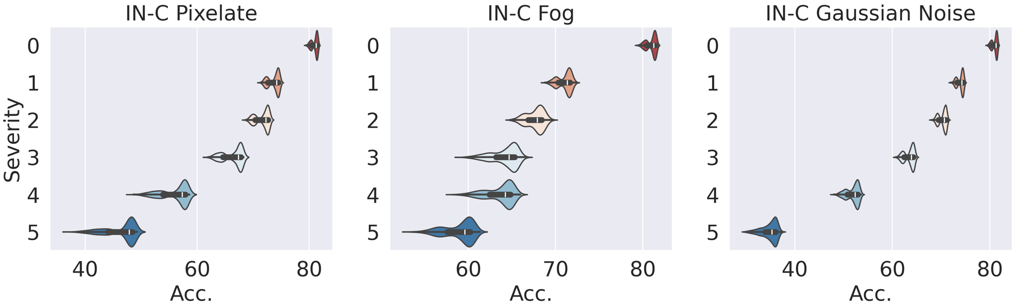}
	% figure caption is below the figure
 \vspace{-0.2cm}
	\caption{Visualization of OoD accuracy range across OoD shift severity. We conduct the analysis on $1,000$ architectures in Autoformer-Small search space within our OoD-NAS-ViT benchmarks. Level $0$ denotes the clean examples. All corruptions can be found in Fig. \ref{fig:accuracy_severity-range_rest}, in Appx. \ref{Sec:Appx_OoDrange}. \textbf{We generally observe that the range of OoD accuracy widens as the severity of the OoD shift increases.}}
 \vspace{-0.5cm}
	\label{fig:OoD-range-severity}       % Give a unique label
\end{figure}

% So far, we establish that ViT OoD and ID accuracy are not very highly correlated. 
In this section, we highlight that ViT architectures considerably impact OoD accuracy. This observation encourages future research to put more focus on ViT architecture research for OoD generalization. 

\textbf{Experimental Setups. }To show how ViT architecture designs impact OoD generalization, we compute the range of OoD accuracy for each search space on an OoD dataset. This range reflects the variation in OoD performance for different architectures within a search space. For example, in Fig.~\ref{fig:OoDshifts_Visualization}, each sub-plot represents the range of OoD accuracy for three different search spaces in our OoD-ViT-NAS benchmark on one OoD dataset. We compute the average OoD accuracy range across all datasets as general statistics. For reference, the range of ID accuracy is also included.

\textbf{Results. }The results are shown in Fig.~\ref{fig:OoDshifts_Visualization}. Additional results can be found in the Appx.\ref{Sec:Appx_OoDrange}. The average range of OoD accuracy across the three search spaces in our benchmark is $3.81\%$/$4.86\%$/$2.74\%$, which is comparable to or even outperforming state-of-the-art (SOTA) training OoD generalization method, such as those based on domain invariant representation learning \cite{bai2024hypo,arjovsky2019invariant}. For example, under a similar OoD setting, the current SOTA \cite{bai2024hypo} shows an improvement of $1.9\%$ OoD accuracy. This observation highlights the significant influence of ViT architectural designs on OoD accuracy. By carefully designing ViT architecture, the OoD accuracy could improve significantly.

We further explore how the severity of the OoD shift affects the range of ViT's OoD accuracy. We conduct similar experimental setups as before, analyzing $1,000$ architectures from the Autoformer-Small search space within our OoD-ViT-NAS benchmark for $1,000$ architectures in Autoformer-Small search space within our benchmark on IN-C.
% We focus on the IN-C dataset because it provides variations in OoD shift severity. 
The results are visualized in Fig.~\ref{fig:OoD-range-severity}. We observe that the range of OoD accuracy widens as the severity of the OoD shift increases. This suggests that under stronger OoD shifts, the architecture design becomes even more critical for OoD generalization.

When visualizing OoD accuracy, we observe a bimodal distribution. We figure out that the embedding dimension, as the primary ViT structural attribute, 
 influences this bimodality. For example, among architectures from the Autoformer-Small search space of our benchmark, most architectures with a lower embedding dimension ($320$) fall within the lower OoD accuracy mode, while those with higher dimensions ($384$ and $448$) tend to reside in the higher accuracy mode. This observation will be further discussed in detail in Sec.~\ref{Sec:Robust_Arch}.

\subsection{Can ID accuracy serve as a good indication for OoD accuracy?} \label{Sec:IDvsOoDAcc}

While existing works \cite{tu2022maxvit,liu2021swin,chen2021autoformer,wang2023prenas,su2022vitas,tang2023elasticvit,gong2022nasvit,ding2021hr} study the impact of ViT architectures to ID accuracy, studies on OoD accuracy are limited. To what extent can we directly apply existing findings of ViT architecture insights for ID to OoD accuracy? To answer this question, we investigate the relationship between ViT ID and ViT OoD accuracy. 
% If they are highly correlated, existing finding for ID acc could be applicable to OoD.

Several studies \cite{Miller2021,Recht2019, Teney2023, Wenzel2022} investigates the relationship between ID and OoD accuracy  for the CNNs model. However, there is no work on such study particularly for ViT. Utilizing our OoD-ViT-NAS benchmark, we provide the first comprehensive study on the relationship between ViT ID and OoD accuracy. Through our investigation, we find that the correlation between ViT ID and ViT OoD accuracy is not very high. This suggests that architectural insights optimized for ViT ID accuracy, as presented in previous work \cite{tu2022maxvit,liu2021swin,chen2021autoformer,wang2023prenas,su2022vitas,tang2023elasticvit,gong2022nasvit,ding2021hr} may not be applicable for ViT OoD generalization.

\begin{figure}[t]
	\centering
	% Use the relevant command to insert your figure file.
	% For example, with the graphicx package use
	\includegraphics[width=.95\linewidth]{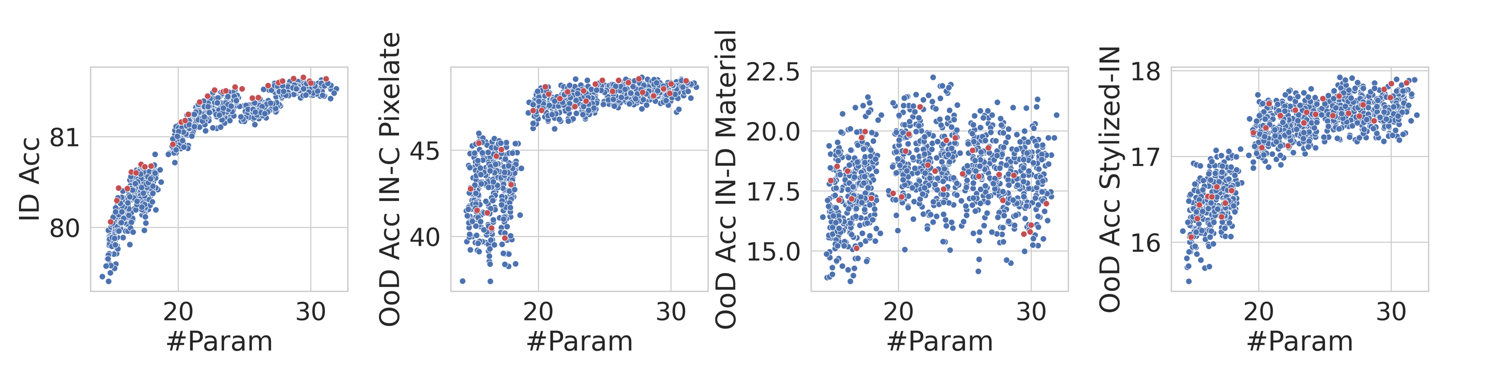}
	% figure caption is below the figure 
 %1.\linewidth]
 \vspace{-0.2cm}
	\caption{Analysis of OoD Generalization Performance of Pareto Architectures for ID accuracy. Blue dots \tikzcircle[bluecolor, fill=bluecolor]{2pt} represent architectures in the search space, while red dots \tikzcircle{2pt} represent the ID Pareto architectures. See Appx. \ref{Sec:Appx_Pareto} for additional results. \textbf{We find that Pareto architectures for ID accuracy generally perform sub-optimally under OoD shift.}}
 \vspace{-0.8cm}
	\label{fig:pareto}       % Give a unique label
\end{figure}

\textbf{Experimental Setup.} Following previous work \cite{Teney2023}, we use Kendall's $\tau$ rank correlation coefficient to compute the correlation between OoD and ID accuracy of all $1,000$ architectures from a search space on one OoD dataset. Our examination comprehensively computes the correlations across all $8$ OoD datasets, $3$ search spaces, and $3,000$ architectures within our OoD-ViT-NAS benchmark. We compute the average correlations across search spaces and datasets as general statistics.

Besides the investigation of the correlation of various architectures, we further study the relationship between ViT ID and ViT OoD accuracy for Pareto architectures, representing the top-performing architectures for a certain model size. As shown in Fig.~\ref{fig:pareto}-a, the red dots \tikzcircle{2pt} represent Pareto architectures for ID accuracy
% , achieving the highest performance within their respective parameter size constraints. 
In this study, we analyze $1,000$ architectures from the Autofomer-Small search space within our OoD-ViT-NAS benchmark. To identify Pareto architectures for ID accuracy, we divide the total parameter budget into $30$ equal intervals and select the architecture with the best ID performance within each interval.

\textbf{Results.} The correlation results are illustrated in Fig.~\ref{fig:analysis-investigation}-a. The individual correlations can be found in the Appx.~\ref{Sec:Appx_IDvsOoD}. We show that the correlation between ID and OoD accuracy is generally not very high. This suggests that current architectural insights based solely on ID accuracy might not effectively translate to OoD generalization.

Among all OoD datasets, the IN-P dataset exhibits the strongest correlation with ID accuracy. This can be attributed to its weaker OoD shift compared to other datasets (see the visualization in Appx.~\ref{Sec:Appx_Benchmark}). As a result, the OoD examples in IN-P are not very different from ID examples, leading to a relatively high correlation between OoD and ID performance. For the remaining seven datasets with stronger OoD shifts, the correlations remain relatively low. 
% Further discussion can be found in the Appx.~\ref{}.

The results of the Pareto architectures analysis are illustrated in Fig.~\ref{fig:pareto}. Additional results can be found in the Appx.~\ref{Sec:Appx_Pareto}. We observe that Pareto architectures for ID accuracy generally perform sub-optimally under the OoD shift.
This observation further supports our previous finding that ID accuracy might not be a very good indicator of OoD accuracy. 

% \vspace{-1.5cm}
\subsection{Explore Training-free NAS for OoD Generalization} \label{Sec:Proxies}

Recently, there has been a new research focus on Training-free NAS, aimed at identifying high-performing architectures without the computational expense of training each candidate. To do so, \cite{zhou2022training,zhou2024training,wei2024auto,ha2024generalizable,jiang2024meco,lee2018snip,wang2020picking} propose zero-cost proxies to predict the performance of candidate architectures in the initialization or the first training iteration, significantly accelerating NAS. While these works focus on ID accuracy, a few attempts have been made in searching for architectures robust against adversarial attacks \cite{ha2024generalizable,hosseini2021dsrna}. However, there is no work to explore Training-free NAS for ViT for OoD generalization.

\textbf{Experimental Setup.} To address this gap, we comprehensively explore the existing $9$ Training-free NAS for OoD generalization on $3,000$ ViT architectures within our OoD-ViT-NAS benchmark. Our study includes common and SOTA Training-free NAS originally proposed for CNNs for ID Acc (Grasp \cite{wang2020picking}, SNIP \cite{lee2018snip}, MeCo \cite{jiang2024meco}), ViTs for ID Acc (DSS \cite{zhou2022training}, AutoProx \cite{wei2024auto}), and CNNs for adversarial robustness (Jacobian \cite{hosseini2021dsrna}, CroZe \cite{ha2024generalizable}). We complement this study on Training-free NAS to our OoD-ViT-NAS benchmark to equip the NAS research community with valuable tools to develop more effective Training-free NAS for OoD generalization.

\begin{table}[t]
	\centering
 \vspace{-0.5cm}
	\caption{Comparison of Kendall $\tau$ ranking correlation between the OoD accuracies and the Training-free NAS proxies values on 8 common large OoD datasets using our OoD-ViT-NAS benchmark. \textbf{Bold} and \underline{underline} stand for the best and second, respectively. We show that existing Training-free NAS's predictability in ViT OoD accuracy is limited, E.g., the very recently proposed Auto-Prox only achieves $0.3303$ correlation. Furthermore, we make the first observation that simple proxies like \#Param or \#Flops outperform other more complex proxies in predicting both ViT OoD/ID accuracy.}
    \renewcommand{\arraystretch}{1.0}
	\resizebox{0.85\textwidth}{!}{%
		\begin{tabular}{@{}lcccc@{}}
			\toprule
            \multicolumn{1}{c}{\multirow{2}{*}{\textbf{Training-free NAS}}} & \multicolumn{2}{c}{\textbf{Originally Proposed For}}    & \multirow{2}{*}{\textbf{Correlation with ID Acc}} & \multirow{2}{*}{\textbf{Correlation with OoD Acc}} \\ \cline{2-3}
            \multicolumn{1}{c}{}                                   & \multicolumn{1}{c}{\textbf{Performance}} & \textbf{Architecture} &                                          &                                           \\ 
			\midrule
			Grasp \cite{wang2020picking}    & \multicolumn{1}{c}{ID Acc}          & CNNs           & 0.1490 $\pm$ 0.1951     & 0.1207 $\pm$ 0.1575 \\
   
			SNIP \cite{lee2018snip}      & \multicolumn{1}{c}{ID Acc}          & CNNs           & 0.3750 $\pm$ 0.3023     & 0.2889 $\pm$ 0.2274\\
   
			MeCo \cite{jiang2024meco}      & \multicolumn{1}{c}{ID Acc}          & CNNs           & 0.1440 $\pm$ 0.2371 & 0.0975 $\pm$ 0.0819\\

			CroZe \cite{ha2024generalizable}  & \multicolumn{1}{c}{Adv Robustness}  & CNNs           & 0.3823 $\pm$ 0.3046  & 0.2951 $\pm$ 0.2223 \\
   
			Jacobian \cite{hosseini2021dsrna} & \multicolumn{1}{c}{Adv Robustness}  & CNNs           & 0.1053 $\pm$ 0.1509 & 0.0841 $\pm$ 0.1232 \\

			DSS \cite{zhou2022training}      & \multicolumn{1}{c}{ID Acc}          & ViTs           & 0.4165 $\pm$ 0.3461 & 0.3421 $\pm$ 0.2365  \\
	
			AutoProx-A \cite{wei2024auto}& \multicolumn{1}{c}{ID Acc}         & ViTs           & 0.4023 $\pm$ 0.3827 & 0.3303 $\pm$ 0.2384  \\
   
            \rowcolor{lightgray}
			\#Param   & -               & -             & \underline{0.4607} $\pm$ \underline{0.3318} & \underline{0.3600} $\pm$ \underline{0.2321} \\
   
			\rowcolor{lightgray}
			\#Flops   & -               & -             & \textbf{0.4705} $\pm$ \textbf{0.3391} & \textbf{0.3537} $\pm$ \textbf{0.2327}  \\
   
			\bottomrule
		\end{tabular}}
  \vspace{-0.5cm}
  \label{tab:proxy}
\end{table}

\textbf{Results.} Our exploration provides several practical insights for designing a Training-free NAS for ViT for OoD generalization. The results of Kendall $\tau$ \cite{kendall1938new} ranking correlation between the OoD accuracy and Training-free NAS proxies on $8$ common large OoD datasets are illustrated in Tab.~\ref{tab:proxy}. The average OoD accuracy is computed across OoD datasets and search spaces. Detailed results can be found in Fig.~\ref{fig:analysis-investigation} and Appx.~\ref{Sec:Appx_Proxies}.

We observe that existing Training-free NAS are largely ineffective in predicting OoD accuracy. Even recent Training-free NAS designed for ViT (i.e., DSS \cite{zhou2022training} and AutoProx-A \cite{wei2024auto} ) or Training-free NAS designed adversarial robustness \cite{ha2024generalizable} struggle with predicting OoD accuracy. 

Surprisingly, simple zero-cost proxies such as \#Param or \#Flops outperform all existing, more complex proxies in predicting both OoD accuracy for ViTs. This finding poses a challenge to the Training-free NAS research community: to devise a Training-free NAS that surpasses \#Params or \#Flops in OoD Acc prediction for ViT.

From Fig.~\ref{fig:analysis-investigation}-b, we observe that all Training-free NAS methods consistently fail to predict IN-D performance. This is due to the IN-D dataset's unique generation process using Stable Diffusion, which creates images labelled with object names and varying nuisances like background, texture, and material variations. Only the most challenging images are retained, resulting in highly difficult examples, such as distorted images and unrealistic object-background placements (see Fig.~\ref{fig:imagenet-d-example}). These examples degrade ViT model performance significantly \cite{zhang2024imagenet} and cause unpredictable behaviour.

Our investigation into ID accuracy for ViTs also reveals a surprising observation. While proposals for Training-free NAS designed for ViTs (i.e., DSS \cite{zhou2022training} and AutoProx-A \cite{wei2024auto}), improve the prediction of ID accuracy compared to counterparts designed for CNNs. Our study marks the first attempt to explore simple Training-free NAS like \#Param or \#Flops. The ID prediction of such simple proxies surprisingly outperforms SOTA Training-free NAS designed for predicting ID accuracy for ViT.

\subsection{ViT Structural Attributes on OoD Generalization: Increasing Embedding Dimension is Generally Helpful} \label{Sec:Robust_Arch}

Our OoD-ViT-NAS benchmark with $3,000$ ViT architectures covers diverse design choices in ViT structural attributes, including embedding dimension (Embed\_Dim), network depth, number of heads (\#Heads), and MLP ratio (MLP\_Ratio), which allows for finding a wide range of ViT with different structures and complexities. 
Utilizing our comprehensive benchmark, we are the first to provide an analysis of the impact of these ViT structural attributes. We investigate which structural attributes in ViTs could lead to better OoD generalization. \textit{Through our analysis, we find that increasing the embedding dimension of a ViT architecture can generally improve OoD generalization.} The additional analysis on another search space \cite{dosovitskiy2020image} further confirms our finding. The details for this additional analysis can be found in Appx. ~\ref{Sec:Analysis on Human-design ViT Search Space}

\begin{figure}[t]
	\centering
	% Use the relevant command to insert your figure file.
	% For example, with the graphicx package use
	\includegraphics[width=0.9\linewidth]{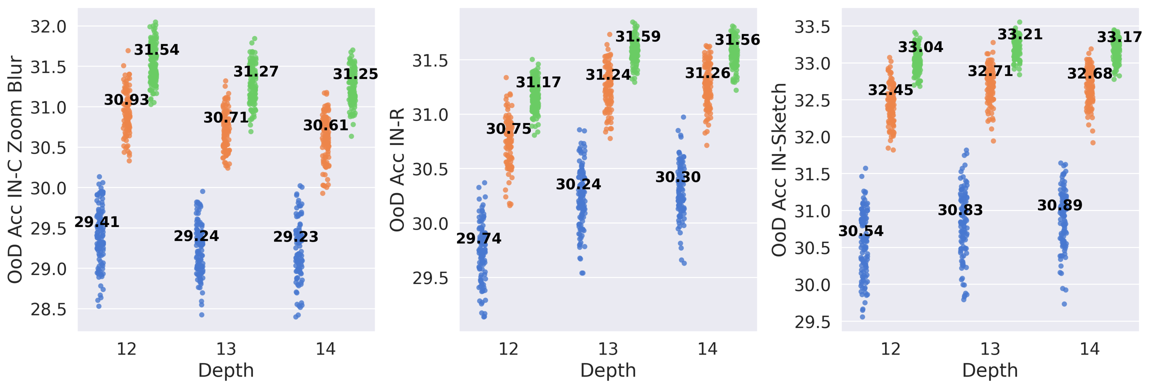}
	% figure caption is below the figure
 \vspace{-0.2cm}
	\caption{The effect of \#Embed\_Dim on robustness generalization of ViTs. The numbers denote the mean OoD accuracy across ViT architectures with specific colour-coded embedding dimensions and depths. The data points with blue \tikzcircle[color320, fill=color320]{2pt}, orange \tikzcircle[color384, fill=color384]{2pt}, and green \tikzcircle[color448, fill=color448]{2pt} colours represent ViT architectures with an embedding dimension of $320$, $384$, and $448$, respectively.  \textbf{Generally, a higher OoD accuracy is obtained when the embedding dimension of ViT architectures increases for most OoD shifts.} See Fig. \ref{fig:embedding-all-small1} and \ref{fig:embedding-all-small2} in Appx. \ref{Sec:Appx_ViTAttribute_Embedding} for additional plots and results on other OoD shifts. }
    \vspace{-0.4cm}
	
	\label{fig:embedding}       % Give a unique label
\end{figure}

\paragraph{Experimental Setup.} To verify the effectiveness of ViT architectural attributes on our OoD generalization benchmark, we present the results from two perspectives: (1) an analysis of rank correlation for our OoD-ViT-NAS benchmark and (2) a comparison of OoD accuracy across different embedding dimensions. To gain insights into the relationship between embedding dimension (Embed\_Dim) and OoD performance, we created the visualizations for all architectures in our OoD-ViT-NAS benchmark. These visualizations compare the average OoD accuracies across different ViT architectures with varying embedding dimensions and depths. Examples of these visualizations are shown in Fig. \ref{fig:embedding}. Additional results can be found in the App.\ref{Sec:Appx_ViTAttribute_Embedding}.

\paragraph{Results.} In Fig. \ref{fig:analysis-investigation}-c, we find that the embedding dimension generally has the highest correlation with OoD accuracy among all ViT architectural attributes. This positive correlation indicates that \textit{Embed\_Dim could play a crucial role in achieving OoD generalization performance.} Our comprehensive OoD-ViT-NAS benchmark sheds light on a previously unknown relationship: the potential impact of embedding dimension (Embed\_Dim) on OoD generalization in ViTs.  This trend holds across most OoD shifts in our benchmark (Fig.~\ref{fig:embedding}), suggesting that among other architectural attributes, the design choice of Embed\_Dim might significantly influence a model's OoD generalization. 

Our experiments yield several intriguing phenomena. Based on Fig. \ref{fig:analysis-investigation}-c, we observe that \textit{network depth has a slight impact on overall OoD generalization performance (correlation: $0.19$).} Also, as shown in Fig. \ref{fig:embedding}, for a given embedding dimension (represented by a distinct colour), we report the mean OoD accuracy, showing how the mean OoD accuracy changes among ViT architectures of varying depths, which aligns with our empirical insight. Fig. \ref{fig:embedding} shows that while increasing depth can be beneficial for improving ViT's OoD generalization in some cases, there exist shallower models that tend to perform better in terms of OoD accuracy compared to those with deeper models. 

It is evident from Fig.~\ref{fig:analysis-investigation}-c, where both the MLP ratio (0.09) and the number of heads ($0.07$) exhibit very low correlation values with overall OoD performance. These findings highlight that increasing the MLP ratio and the number of heads may not substantially enhance a model’s robustness to OoD data. Due to space constraints, we defer additional experiments to the Appx.~\ref{Sec:Appx_ViTAttribute}, showing that the network depth, MLP ratio, and \#Heads might have non-obvious impacts on OoD generalization.

\begin{figure}[t]
\centering
 \vspace{-0.3cm}
	\includegraphics[width=0.9\linewidth]{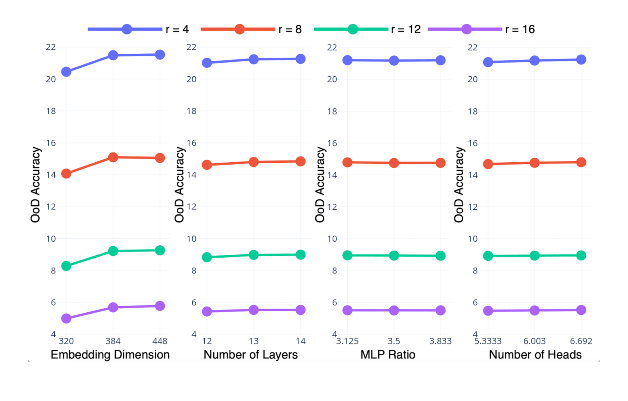}
 \vspace{-0.8cm}
	\caption{Following setting in \citep{bai2022improving,wang2020high,gavrikov2024can}, the ViTs which were trained on original ID data, are now tested on high frequency components (HFC) of OoD samples, with r as the radius for frequency filtering. The higher the OoD accuracy, the more HFC learned in the model.}  
 \vspace{-0.7cm}
	\label{fig:frequency_analysis}       
\end{figure}

\textbf{Increasing Embedding Dimension help ViT learn more high-frequency patterns, leading to improve OOD generalization.} In this section, we design a frequency study to understand our finding: why increasing ViT Embedding Dimension can generally improve ViT’s OOD generalization. In the literature, the models obtain higher performance on preserving High-Frequency-Component (HFC) samples tend to learn more HFC \citep{bai2022improving,shao2021adversarial}. By learning more HFC, the models improve OOD generalization \citep{bai2022improving,gavrikov2024can}. Our hypothesis is that Increasing embedding dimension helps ViTs learn more HFC resulting in improving OOD generalization. We adapt the experiment from \citep{bai2022improving} to verify our hypothesis. The details on this experimental setup can be found in the Appx.~\ref{Sec:App_Freq_Analysis}. In a nutshell, we filter HFC by hyper-parameter radius r, where the higher the r, the lesser HFC. As shown in Fig.~\ref{fig:frequency_analysis}, we observe that when increasing Embedding Dimension, the performances obtained on filtering-HFC samples are improved. This observation holds true across setups varying radius r, supporting that increasing Embedding Dimension helps ViT learn more HFC. In contrast, increasing other ViT structural attributes does not help improve ViT learn more HFC. 

\paragraph{Robust ViT architectures designed by our finding.} Our study provides significant insights for guiding the design of ViT architectures. Specifically, among ViT structural attributes, increasing embedding dimension can generally improve OoD generalisation of ViT architectures. Our insight leads to a simple method which can achieve ViT architectures that can outperform well-established human-designed. We demonstrate the superiority of ViT based on our insights in Tab.~\ref{tab:application}. Scaling up ViT architecture (e.g., from ViT-B-32 to ViT-L-32) by humans typically involves compound scaling of various ViT structural attributes. However, our findings suggest that not all ViT structural attributes need to be increased to benefit OoD generalisation. Among these attributes, increasing the embedding dimension is the most crucial factor for improving OoD generalisation. By only increasing the embedding dimension, ours ViT architectures (e.g., Increasing embedding dimension of ViT-B-32) are significantly more efficient and outperform compound scaling architectures (e.g., ViT-L-32).

\begin{table}[]
\renewcommand{\arraystretch}{1.5}
\centering
\caption{%Comparison of ViT architectures designed based on our insights and those designed by humans.
Comparison of ViT architectures designed based on our insights (\textbf{Embedding Dimension}) and well-established ViT designed by humans in \cite{liu2021swin,dosovitskiy2020image}. 
% By only increasing the embedding dimension, our ViT architectures are significantly more efficient and outperform compound scaling architectures.
% Note the superiority of ViT based on our insights. Scaling up ViT architecture (e.g., from ViT-B-32 to ViT-L-32) by humans typically involves compound scaling of various ViT structural attributes. However, our findings suggest that not all ViT structural attributes need to be increased to benefit OoD generalisation. Among these attributes, increasing the embedding dimension is the most crucial factor for improving OoD generalisation.
}
\resizebox{.9\textwidth}{!}{%
\begin{tabular}{cccccccccc}
\hline
\textbf{Architecture} & \textbf{\makecell{Embed \\ Dim}} & \textbf{Depth} & \textbf{\#Head} & \textbf{\makecell{MLP \\ Ratio}} & \textbf{\makecell{Latency \\ (ms)}} & \textbf{\makecell{\#Param \\ (M)}} & \textbf{\makecell{IN-R \\ OoD Acc}} & \textbf{$\Delta$ wrt Latency $\uparrow$} & \textbf{$\Delta$ wrt \#Param $\uparrow$} \\ \hline
ViT-B-32                               & 768                                           & 12                              & 12                               & 4.0                                 & 105.00                                 & 87.53                                 & 41.58                                  & -                                       & -                                       \\
Ours    & 840                                           & 12                              & 12                               & 4.0                                 & 113.32                                 & 98.64                                 & \textbf{48.28}                                  & \textbf{0.8054}                         & \textbf{0.6032}                         \\
ViT-L-32 \cite{dosovitskiy2020image}                               & 1024                                          & 16                              & 24                               & 4.0                                 & 258.83                                 & 305.61                                & 44.33                                  & 0.0179                                  & 0.0126                                  \\ \hline
Swin-T                                 & 96                                            & 12                              & 32                               & 4.0                                 & 100.31                                 & 28.29                                 & 46.22                                  & -                                       & -                                       \\
Ours      & 128                                           & 12                              & 32                               & 4.0                                 & 165.80                                 & 49.91                                 & \textbf{48.28}                                  & \textbf{0.0315}                         & \textbf{0.0954}                         \\
Swin-S \cite{liu2021swin}                               & 96                                            & 24                              & 32                               & 4.0                                 & 184.48                                 & 49.60                                 & 47.77                                  & 0.0184                                  & 0.0725                                  \\ \hline
\end{tabular}%
}
\label{tab:application}
\end{table}

%% file: sections/conclusion.tex
\section{Conclusion}

In this work, we introduce OoD-ViT-NAS, the first comprehensive benchmark for NAS on OoD generalization of ViT architectures. Using this benchmark, we conduct a comprehensive investigation on OoD generalization for ViT. Firstly, we show that ViT architecture design significantly impacts OoD accuracy. Secondly, we show that the architectural findings from existing works for ID performance could not apply to OoD generalization due to the low correlation between ID and OoD accuracy. Thirdly, we conduct the first study of NAS for ViT’s OoD generalization and show that existing Training-free NAS methods struggle with OoD prediction. Surprisingly, simple proxies like \#Param or \#Flops outperform other complex Training-free NAS. Finally, we conduct the first study on the impact of  ViT architectural attributes on OoD generalization. Our study reveals that increasing a ViT architecture's embedding dimensions can generally improve OoD generalization. We believe our benchmark OoD-ViT-NAS and comprehensive analysis will catalyze and streamline future research on understanding how ViT architecture design influences OoD generalization.

\noindent {\bf Acknowledgement.} This research is supported by the National Research Foundation, Singapore under its AI Singapore Programmes (AISG Award No.: AISG2-TC-2022-007); The Agency for Science, Technology and Research (A*STAR) under its MTC Programmatic Funds (Grant No. M23L7b0021). This research is supported by the National Research Foundation, Singapore and Infocomm Media Development Authority under its Trust Tech Funding Initiative. Any opinions, findings and conclusions or recommendations expressed in this material are those of the author(s) and do not reflect the views of National Research Foundation, Singapore and Infocomm Media Development Authority.

%% file: sections/appendix.tex
\newpage
% {\Large Appendix}

\section{Appendix Overview} \label{Sec:Appx_Overview}

% The PyTorch code and our benchmark can be accessed at the following anonymous links:

% \begin{itemize}
%     \item Code for constructing our OoD-ViT-NAS benchmark: \href{https://anonymous.4open.science/r/OoD-ViT-NAS-B507/}{\color{red} available here}
%     \item Code for exploring Training-free NAS for ViT OoD generalization: \href{https://anonymous.4open.science/r/OoD-ViT-NAS-B507/}{\color{red} available here}
%     \item Benchmark: \href{https://drive.google.com/drive/folders/19nb6d2DttDCLhWNY117M8nFn7zxb2LH2?usp=sharing}{\color{red} available here}
% \end{itemize}

This appendix provides supplementary information that are not included in the main paper due to space limitations. 
% The appendix consists of:

\unhidefromtoc
\tableofcontents
\clearpage
\renewcommand\thefigure{\thesection.\arabic{figure}}
\renewcommand\theHfigure{\thesection.\arabic{figure}}
\renewcommand\thetable{\thesection.\arabic{table}}  
\renewcommand\theHtable{\thesection.\arabic{table}}  
\setcounter{figure}{0} 
\setcounter{table}{0}

\input{sections/limitation}
\input{appx_sections/observation-on-vit-search-space}
\input{appx_sections/Detailed-description-of-Frequency-Analysis}

\input{appx_sections/benchmark}

\newpage
\input{appx_sections/benchmark_usage}

\input{appx_sections/ood_range}
\input{appx_sections/idvsood}
\newpage
\input{appx_sections/pareto}

\newpage

\input{appx_sections/proxy}
\newpage

\input{appx_sections/vit_attribute_embedding}

\newpage

\input{appx_sections/vit_attribute}

\newpage

\input{appx_sections/reproducibility}
\newpage

%% file: sections/limitation.tex
\section{Limitations and Broader Impact} \label{Sec:Limitation}

Given the extensive set of experiments presented in this work, the evaluation of training-free NAS proxies is significantly dependent on the initial robust benchmarks, which can be costly and resource intensive to create in the first place.

As this work studies the robustness of ViT architectures to OoD shift, we could demonstrate that carefully designed ViT architectures can significantly enhance OoD generalization. Our approach focuses on evaluating training-free NAS for ViT architectures, offering valuable insights that can complement research exploring the effects of robust architectural design. Finally, we publicly release our OoD-NAS-ViT and code base for future research.

%% file: appx_sections/observation-on-vit-search-space.tex
\section{Analysis on Human-design ViT Search Space} \label{Sec:Analysis on Human-design ViT Search Space} 

Our findings on embedding dimensions are derived from analysing ViT architectures sampled through AutoFormer. To further validate these results, we also investigate the impact of ViT structural attributes on OoD generalisation within the human-designed ViT search space \citep{dosovitskiy2020image}.

\begin{table}[h]
\renewcommand{\arraystretch}{3.0}
\caption{We conduct additional experiments on human design ViT to further confirm our main findings that, among ViT structural attributes, embedding dimension is the most important ViT structural attribute to OoD generalisation. We train each architecture from on IN-100 and evaluate on IN-R. \textbf{The trade-off between OoD Acc and computational metrics (i.e., Latency and \#Param) is quantifies by $\Delta$, which is the ratio of increase in OoD Acc and increase in computational metrics.} Higher $\Delta$ is better. Note that increasing \#Head remains the same \#Param but increases Latency in ViT setup \citep{dosovitskiy2020image}.} 
\resizebox{1.\textwidth}{!}{%
\begin{tabular}{cccccccccc}
\hline
 \multirow{2}{*}{\textbf{Architecture}} & \multicolumn{4}{c}{\textbf{Configuration}}                                 & \multirow{2}{*}{\textbf{\begin{tabular}[c]{@{}c@{}}Latency \\ (ms)\end{tabular}}} & \multirow{2}{*}{\textbf{\begin{tabular}[c]{@{}c@{}}\#Param \\ (M)\end{tabular}}} & \multirow{2}{*}{\textbf{OoD Acc}} & \multirow{2}{*}{\textbf{$\Delta$ wrt Latency $\uparrow$}} & \multirow{2}{*}{\textbf{$\Delta$ wrt \#Param $\uparrow$}} \\ \cline{2-5}
 
                          & \textbf{\makecell{Embed \\ Dim}} & \textbf{\#Head} & \textbf{Depth} & \textbf{\makecell{MLP \\ Ratio}} &                                                                                   &                                                                                  &                                        &                                         &                                         \\ \hline
                          
 Vanila                                 & 768                & 12              & 12             & 3072               & 105.00                                                                            & 87.53                                                                            & 41.58                                  & -                                       & -                                       \\
 
 \makecell{Increased \\ Embed-Dim}                    & { 840}          & 12              & 12             & 3072               & 113.32                                                                            & 98.64                                                                            & \textbf{48.28}                                  & \textbf{0.8053}                         & \textbf{0.6032}                         \\

 \makecell{Increased \\ \#Head}                       & 768                & { 128}       & 12             & 3072               & 120.44                                                                            & 87.53                                                                            & 43.47                                  & 0.1224                                  & \textbf{-}                              \\

\makecell{Increased \\ Depth}                        & 768                & 12              & {16}       & 3072               & 137.36                                                                            & 115.88                                                                           & 37.80                                  & -0.1168                                  & -0.1333                                  \\

\makecell{Increased \\ MLP-Ratio}                    & 768                & 12              & 12             & { 3840}         & 120.05                                                                            & 101.70                                                                           & 42.61                                  & 0.0685                                 & 0.0728                                  \\ \hline
\end{tabular}%
}

\label{tab:debias_vit_attribute}
\end{table}

\textit{Experimental setup}. We begin with the vanilla ViT-B-32 architecture \citep{dosovitskiy2020image}, varying each structural attribute independently. For the altered ViT architectures, we increase these attributes to ensure that the capacities of the altered models remain comparable. Each architecture is trained on IN-100 and evaluated on IN-R.

\textit{Results.} As shown in Tab.~\ref{tab:debias_vit_attribute}, the results further confirms our findings that embedding dimension is the most important ViT structural attribute to OoD generalisation.

%% file: appx_sections/Detailed-description-of-Frequency-Analysis.tex
\section{Detailed description of the Frequency Analysis Setup} \label{Sec:App_Freq_Analysis}

We adapt the experiment from \citep{bai2022improving} to verify our hypothesis. We quantify how the amount of HFC learnt in the ViTs changes if the embedding dimension of ViTs changes. Particularly, we first filter HFC from testing images of IN-R following \citep{bai2022improving,gavrikov2024can,wang2020high} then evaluate the performance of 1000 ViTs in our search space on IN-R with filtering data points. Following \citep{wang2020high} to generate HFC-preserving images, we first convert original images to FFT images. Then, we filter HFC by hyper-parameter radius r (r is set to 4, 8, 12, 16 in our experiments). In a nutshell, the higher the r, the lesser HFC.

%% file: appx_sections/benchmark.tex
\section{The details for our benchmark OoD-ViT-NAS} \label{Sec:Appx_Benchmark}

\subsection{Detailed description of Autoformer Search Space} \label{Sec:Appx_Benchmark_Searchspace}

We strictly follow the search space in Autoformer \cite{chen2021autoformer} to construct our OoD-ViT-NAS. The variable factors of basic transformer blocks include the embedding dimension, Q-K-V dimension, number of heads, MLP ratio, and network depth. The detailed search spaces are illustrated in Tab.~\ref{tab:search_space_autoformer}

\begin{table}[!htb]
\renewcommand{\arraystretch}{1.4}
	\centering
	\caption{Detailed search space used to construct our OoD-ViT-NAS. We strictly follow the search space in Autoformer \cite{chen2021autoformer}.}
	\resizebox{1.0\textwidth}{!}{%
		\begin{tabular}{ccccccc}
            \hline
                                                     &               & \textbf{Embedding Dim} & \textbf{Q-K-V Dim} & \textbf{MLP Ratio} & \textbf{Head Num} & \textbf{Depth Num} \\ \hline
            \multirow{3}{*}{\textbf{Supernet-Tiny}}  & \textbf{Max}  & 192                    & 192                & 3.5                & 3                 & 12                 \\
                                                     & \textbf{Min}  & 240                    & 256                & 4                  & 4                 & 14                 \\
                                                     & \textbf{Step} & 24                     & 64                 & 0.5                & 1                 & 1                  \\ \hline
            \multirow{3}{*}{\textbf{Supernet-Small}} & \textbf{Max}  & 320                    & 320                & 3                  & 5                 & 12                 \\
                                                     & \textbf{Min}  & 448                    & 448                & 4                  & 7                 & 14                 \\
                                                     & \textbf{Step} & 64                     & 64                 & 0.5                & 1                 & 1                  \\ \hline
            \multirow{3}{*}{\textbf{Supernet-Base}}  & \textbf{Max}  & 528                    & 512                & 3                  & 8                 & 14                 \\
                                                     & \textbf{Min}  & 624                    & 640                & 4                  & 10                & 16                 \\
                                                     & \textbf{Step} & 48                     & 64                 & 0.5                & 1                 & 1                  \\ \hline
            \end{tabular}}\label{tab:search_space_autoformer}
\end{table}

\subsection{Detailed description of OoD Datasets} \label{Sec:Appx_Benchmark_Dataset}

\begin{figure}[H]
	\centering
	\includegraphics[width=1.\linewidth]{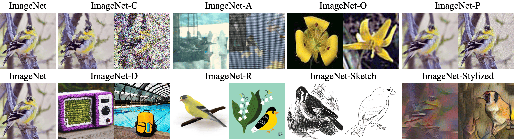}
	\caption{Visualization of different OoD shifts across 8 datasets used to construct our OoD-ViT-NAS benchmark. The description of each dataset can be found in Sec.~\ref{Sec:Appx_Benchmark}.}
	\label{fig:data_visualization}
\end{figure}

To construct OoD-ViT-NAS benchmark, we evaluate 3,000 architectures within our benchmark on common and most SOTA OoD datasets, including:

\begin{itemize}
    \item \textit{ImageNet-1k} \cite{fei2009imagenet}: This is a large and common image dataset widely used in computer vision research. It contains over 1.3 million labeled high-resolution images belonging to 1,000 different object categories (classes). Each image is labeled with a class (e.g., "cat", "airplane", "chair").

    \item \textit{ImageNet-C} \cite{hendrycks2019benchmarking}: 
    This dataset builds upon the original ImageNet test set by applied algorithmically corruptions. These corruptions simulate real-world factors that can deviate data from the training set, such as blur, noise, digital, and weather effects. ImageNet-C offers a comprehensive OoD scenarios with 15 different corruption types, each with 5 severity levels, resulting in a total of 75 unique OoD setups.

    \item \textit{ImageNet-P} \cite{hendrycks2019benchmarking}: 
    This benchmark is constructed similarly to Imagnet-C. The difference is that ImageNet-P utilizes perturbation sequences generated from each ImageNet validation image.  

    \item \textit{ImageNet-A} \cite{hendrycks2021natural}: 
    This dataset is a real-world OoD scenarios by leveraging adversarially filtered images that are highly likely to fool current image classifiers. ImageNet-A selects a 200 classes out of 1,000 classes from ImageNet-1K so that errors among these 200 classes would be considered egregious \cite{fei2009imagenet}. 

    \item \textit{ImageNet-O} \cite{hendrycks2021natural}: 
    Similar to ImageNet-A, this dataset includes adversarially filtered examples specifically designed to challenge OoD detectors trained on ImageNet. ImageNet-O selects a 200 classes out of 1,000 classes from ImageNet-1K.

    \item \textit{ImageNet-R} \cite{hendrycks2021many}: This dataset is a rendition of ImageNet, containing images with manipulated textures and local image statistics.

    \item \textit{ImageNet-Sketch} \cite{wang2019learning}: 
    This dataset introduces a unique OoD challenge by providing black-and-white sketch images corresponding to the ImageNet-1K test set. This significant divergence in visual representation tests a model's ability to generalize beyond photographic data.

    \item \textit{Stylized ImageNet} \cite{geirhos2018imagenet}: 
    This dataset consists of a stylized version of ImageNet generated through techniques like AdaIN \cite{huang2017arbitrary} style transfer, resulting in variations like greyscale, silhouettes, and edges. This dataset assesses a model's ability to handle data with different artistic interpretations.
    
    \item \textit{ImageNet-D} \cite{zhang2024imagenet}: This dataset is a variation of ImageNet generated through diffusion models. These models aim at creating images with rich diversity in backgrounds, textures, and materials. 
\end{itemize}

\begin{figure}[H]
	\centering
	\includegraphics[width=\linewidth]{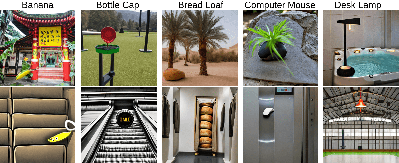}
	\caption{Examples from ImageNet-D \cite{zhang2024imagenet}. These examples are generated by Stable Diffusion \cite{rombach2022high} and only hard examples are kept. These examples could be distorted or unrealistic in object-background placements.}
	\label{fig:imagenet-d-example}
\end{figure}

\begin{figure}[H]
	\centering
	\includegraphics[width=.75\linewidth]{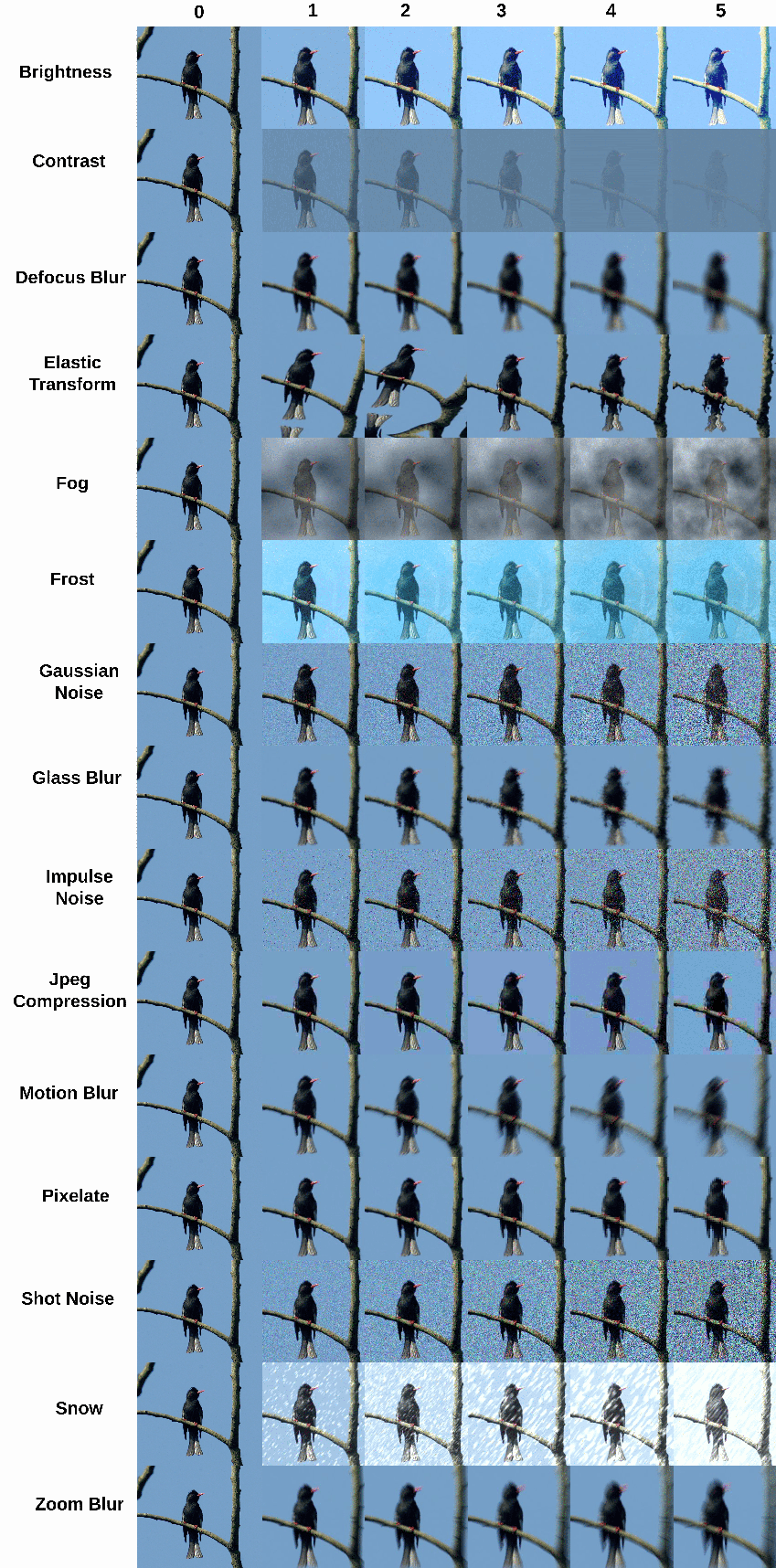}
	\caption{Visualization of different corruptions and 5 different OoD shift severity for ImageNet-C \cite{hendrycks2019benchmarking}. We note that level 0 means clean examples from ImageNet \cite{fei2009imagenet}}
	\label{fig:data_visualization_inc}
\end{figure}

We visualize a few examples of OoD datasets in Fig.~\ref{fig:data_visualization}. For ImageNet-C, we provide the visualization of different OoD shift severity in Fig.~\ref{fig:data_visualization_inc}. 

\newpage

%% file: appx_sections/benchmark_usage.tex
\section{Overview of using the OoD-NAS-ViT benchmark} \label{Sec:Appx_Benchmark_usage}

Our OoD-NAS-ViT benchmark consists of multiple \colorbox{SkyBlue}{json} individual files. Each file includes the evaluation of one search space (Autoformer-Tiny/Small/Base \cite{chen2021autoformer}) on 8 most common and state-of-the-art (SOTA) OoD datasets: ImageNet-C \cite{hendrycks2019benchmarking}, ImageNet-A \cite{hendrycks2021natural}, ImageNet-O \cite{hendrycks2021natural}, ImageNet-P \cite{hendrycks2019benchmarking}, ImageNet-D \cite{zhang2024imagenet}, ImageNet-R \cite{hendrycks2021many}, ImageNet-Sketch \cite{wang2019learning}, and Stylized ImageNet \cite{geirhos2018imagenet} (Sec.~\ref{Sec:Benchmark}).

\begin{figure}[H]
	\centering
	\includegraphics[width=1.0\linewidth]{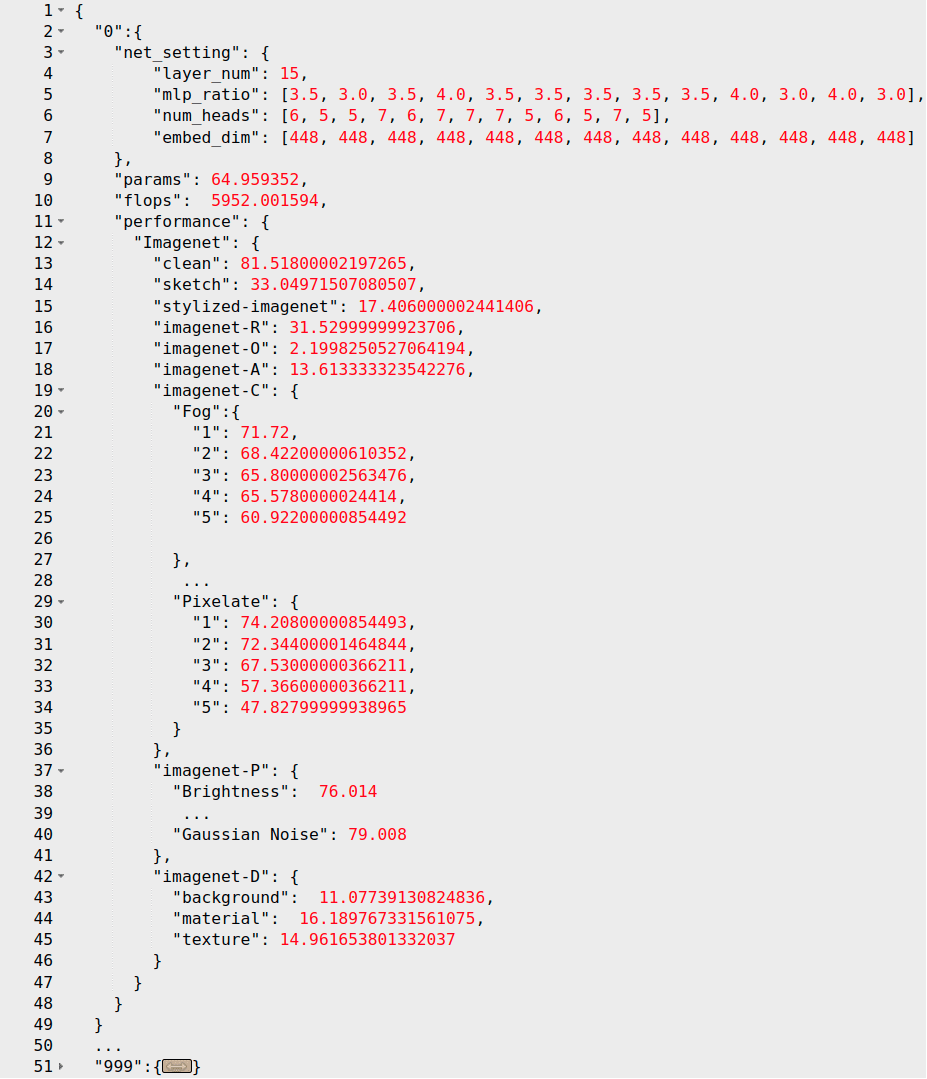}
	\caption{The structure of the merged \colorbox{SkyBlue}{json} file for our OoD-ViT-NAS benchmark for Autoformer-Small search space. The structures of the merged \colorbox{SkyBlue}{json} files for Autoformer-Tiny/Base are similars.}
	\label{fig:benchmark_structure}
\end{figure}

We further provide a merged \colorbox{SkyBlue}{json} file for each search space. The structure of each merged \colorbox{SkyBlue}{json} is illustrated as in Fig.~\ref{fig:benchmark_structure}. Specifically, each merged \colorbox{SkyBlue}{json} file includes 1,000 ViT architectures denoted by the index. Each architecture consists of:

\begin{itemize}
    \item {\color{mygreen} net\_setting}: ViT structural attributes information:
        \begin{itemize}
            \item {\color{mygreen} layer\_num}: network depth
            \item {\color{mygreen} mlp\_ration}: MLP ration each layer, can be varied among layers
            \item {\color{mygreen} num\_heads}: number of attention heads each layer, can be varied among layers
            \item {\color{mygreen} embed\_dim}: embedding dimension heads each layer, fixed among layers
        \end{itemize}
    \item {\color{mygreen} params}: number of parameters
    \item {\color{mygreen} flops}: number of flops
    \item {\color{mygreen} performance}: performance on different dataset:
    
        {\color{mygreen} Imagenet}: performance on IN-based datasets:
                \begin{itemize}
                    \item {\color{mygreen} clean}: performance on IN 
                    \item {\color{mygreen} sketch}: performance on IN 
                    \item {\color{mygreen} stylized-imagenet}: performance on IN 
                    \item {\color{mygreen} imagenet-R}: performance on IN 
                    \item {\color{mygreen} imagenet-O}: performance on IN 
                    \item {\color{mygreen} imagenet-A}: performance on IN-A 
                    \item {\color{mygreen} corruption}: performance on IN-C 
                        \begin{itemize}
                            \item {\color{mygreen} Fog}: performance on one out of 15 corruption in IN-C: Fog, Gaussian Noise, Fog, Snow, Elastic Transform, Jpeg Compression, Frost, Motion Blur, Brightness, Defocus Blur, Glass Blur, Impulse Noise, Shot Noise, Zoom Blur, Constrast, Pixelate
                                \begin{itemize}
                                    \item {\color{mygreen} 1}: performance at OoD shift severity level 1. There are total 5 level of OoD shift severity for each corruption in IN-C
                                \end{itemize}
                        \end{itemize}
                    \item {\color{mygreen} corruption\_P}: performance on IN-P 
                        \begin{itemize}
                            \item {\color{mygreen} Brightness}: performance on one out of 10 corruption in IN-P: Brightness, Motion Blur, Rotate, Scale, Shot Noise, Snow, Tilt, Translate, Zoom Blur, Gaussian Noise
                        \end{itemize}
                    \item {\color{mygreen} imagenet-D}: performance on IN-D 
                        \begin{itemize}
                            \item {\color{mygreen} background}: performance on one out of 3 nuisances in IN-D: background, material, texture
                        \end{itemize}
                \end{itemize}
       
\end{itemize}

With our the provided a merged \colorbox{SkyBlue}{json} file for each search space, we can easily retrieve the OoD shift performance of various ViT architectures and their OoD performance on 8 prevelent and SOTA OoD datasets.

%% file: appx_sections/ood_range.tex
\section{Additional results on the analysis of OoD accuracy range} \label{Sec:Appx_OoDrange}

\begin{figure}[H]
	\centering
	% Use the relevant command to insert your figure file.
	% For example, with the graphicx package use
 \includegraphics[width=1.\linewidth]{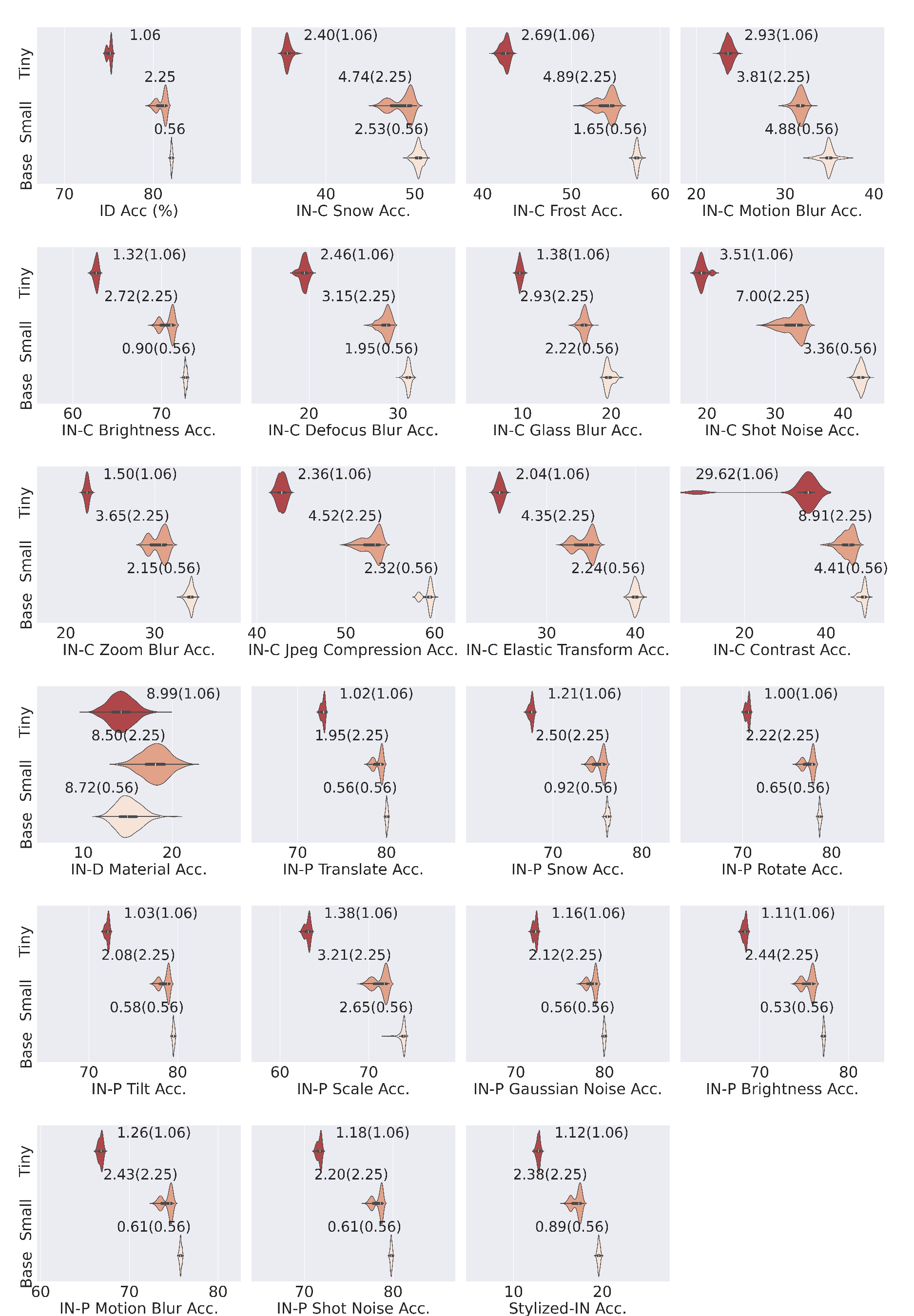}
	% figure caption is below the figure
	\caption{As in Figure \ref{fig:OoDshifts_Visualization}, our analysis on the OoD accuracy range highlights the significant influence of ViT
architectural designs on OoD accuracy. The numbers within each violin plot for each
sub-figures (e.g., IN-D Material 8.99 (1.06), 8.50 (2.25), and 8.72 (0.56)) denote the corresponding OoD(ID)
accuracy range of architectures sampled from AutoFormer-Tiny/Small/Base search space, respectively. }
	\label{fig:accuracy_range_rest}       % Give a unique label
\end{figure}

In the main paper, we visualize $11$ OoD accuracy ranges and ID accuracy range for reference. In this Appx. section, we provide the visualization of the remaining OoD accuracy ranges. The results are illustrated in Fig.~\ref{fig:accuracy_range_rest}. Our observation on other OoD accuracy ranges are generally consistent with our findings in Sec.~\ref{sec:ood_accuracy_range}.

\begin{figure}[h]
	\centering
	% Use the relevant command to insert your figure file.
	% For example, with the graphicx package use
 \includegraphics[width=1.\linewidth]{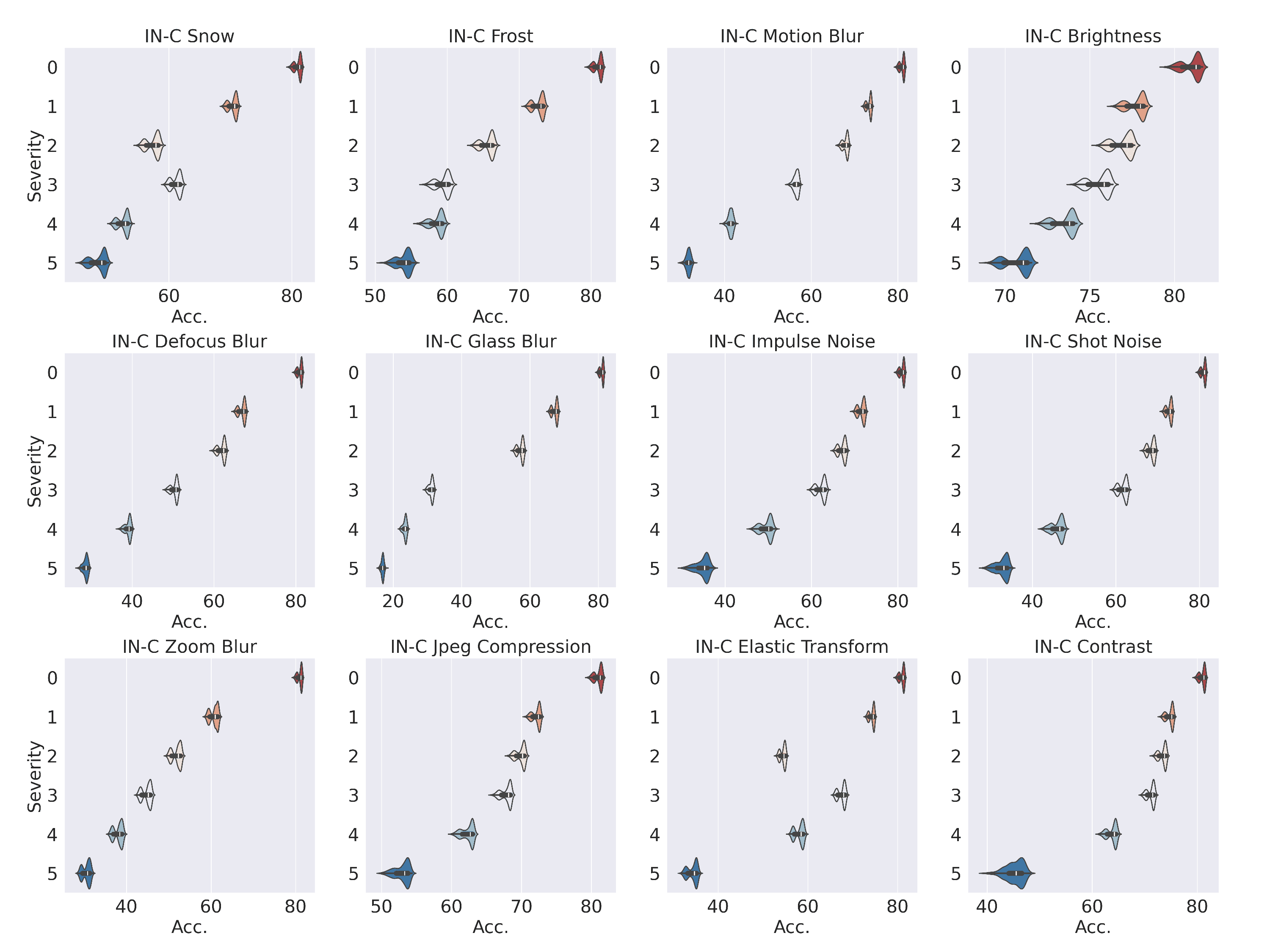}
	% figure caption is below the figure
	\caption{ Visualization of OoD accuracy range across IN-C OoD shift severity. The experiments are conducted on $1,000$ architectures in Autoformer-Small search space within our OoD-NAS-ViT benchmarks. Level $0$ denotes the clean examples. We generally observe that the range of OoD accuracy widens as the severity of the OoD shift increases. }
	\label{fig:accuracy_severity-range_rest}       % Give a unique label
\end{figure}

%% file: appx_sections/idvsood.tex
\section{Additional results on the analysis of the correlation between ID and OoD accuracy} \label{Sec:Appx_IDvsOoD}

\begin{figure}[h]
	\centering
	% Use the relevant command to insert your figure file.
	% For example, with the graphicx package use
 \includegraphics[width=1.\linewidth]{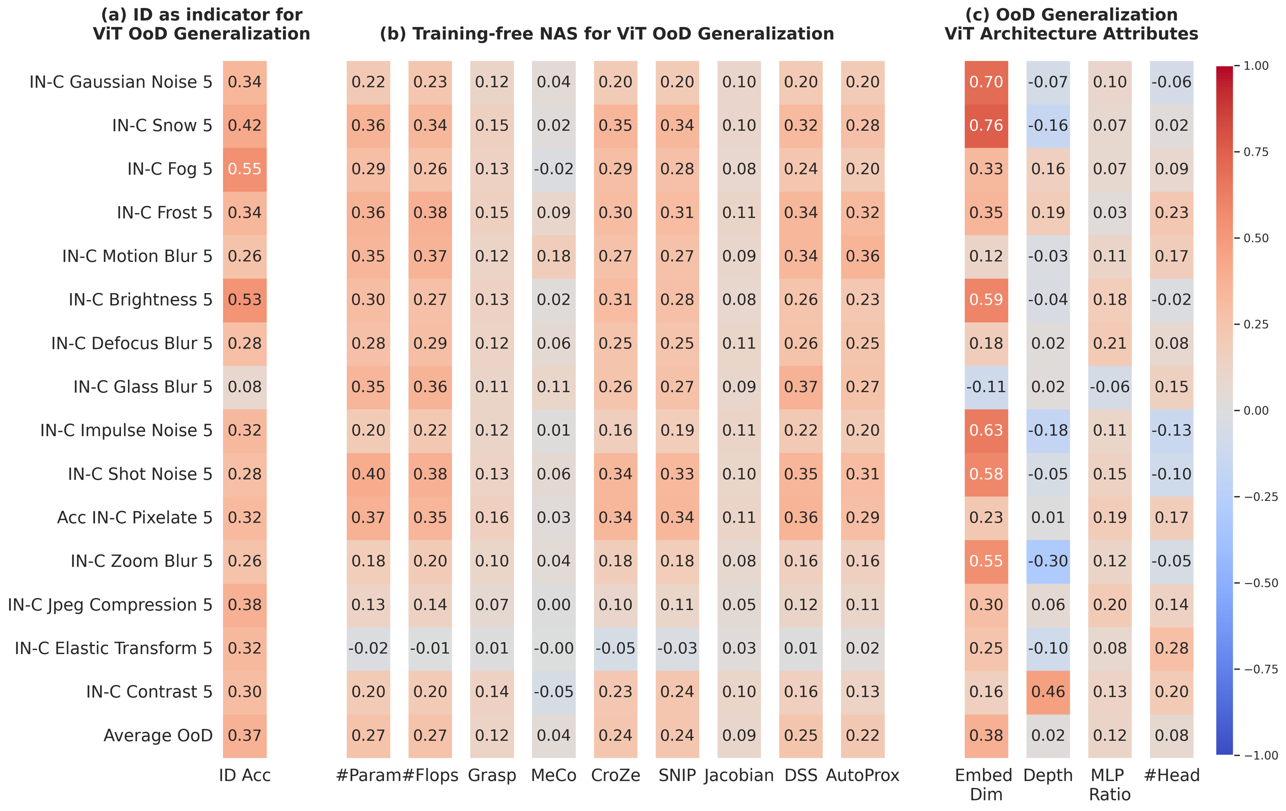}
	% figure caption is below the figure
	\caption{Kendall $\tau$ rank correlation coefficient between ID and OoD accuracies computed on all 3000 architectures in our OoD-ViT-NAS benchmark. Measurements are computed on different corruptions of IN-C. }
	\label{fig:correlation-inc}       % Give a unique label
\end{figure}

\begin{figure}[h]
	\centering
	% Use the relevant command to insert your figure file.
	% For example, with the graphicx package use
 \includegraphics[width=1.\linewidth]{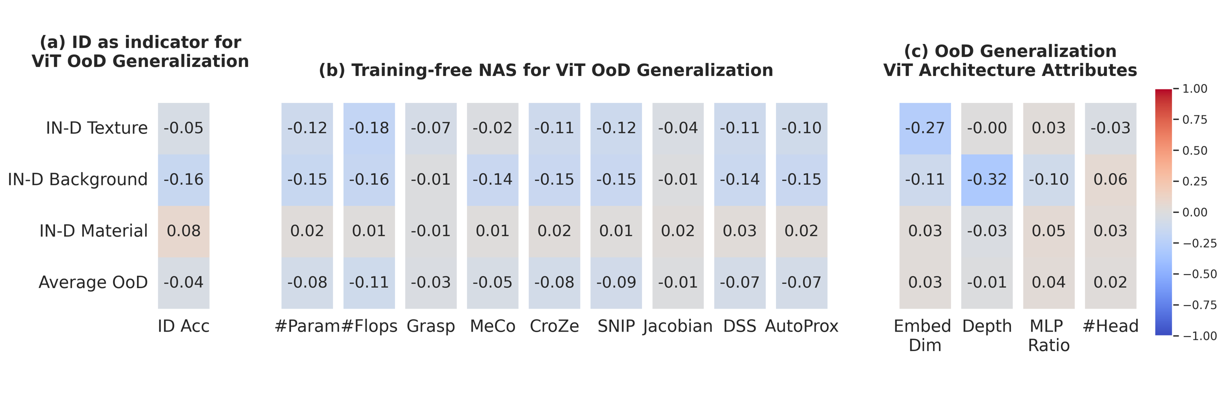}
	% figure caption is below the figure
	\caption{Kendall $\tau$ rank correlation coefficient between ID and OoD accuracies computed on all 3000 architectures in our OoD-ViT-NAS benchmark. Measurements are computed on various IN-D OoD shifts. }
	\label{fig:correlation-ind}       % Give a unique label
\end{figure}

\begin{figure}[h]
	\centering
	% Use the relevant command to insert your figure file.
	% For example, with the graphicx package use
 \includegraphics[width=1.\linewidth]{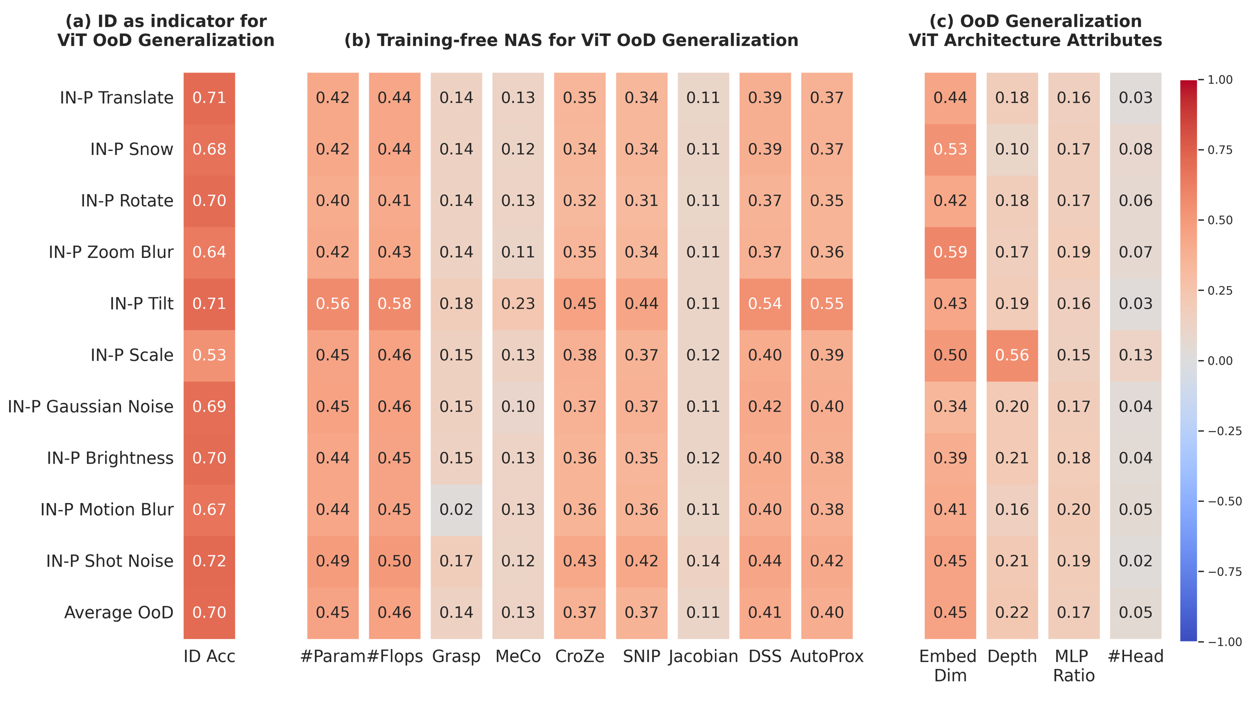}
	% figure caption is below the figure
	\caption{Kendall $\tau$ rank correlation coefficient between ID and OoD accuracies computed on all 3000 architectures in our OoD-ViT-NAS benchmark. Measurements are computed on various IN-P OoD shifts. }
	\label{fig:correlation-inp}       % Give a unique label
\end{figure}

In the main paper, we provide the Kendall $\tau$ correlation between ID accuracy and 8 OoD datasets. For some OoD datasets with different OoD shift types, we average the correlation with ID accuracy of different OoD shift types to obtain average correlation with ID accuracy for that OoD dataset. In this Appx. section, we provide the detailed correlations with ID accuracy of each OoD shifts in such OoD data. Specifically, we provide the detailed correlation for IN-C, IN-D, and IN-P in Fig.~\ref{fig:correlation-inc}, Fig.~\ref{fig:correlation-ind}, and Fig.~\ref{fig:correlation-inp}, respectively.

\begin{figure}[h]
	\centering
	% Use the relevant command to insert your figure file.
	% For example, with the graphicx package use
 \includegraphics[width=1.\linewidth]{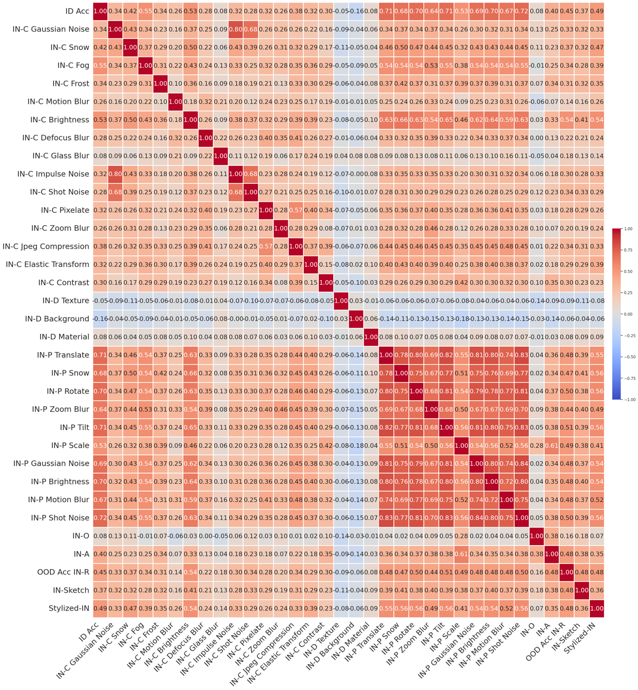}
	% figure caption is below the figure
	\caption{Kendall rank correlation coefficient between ID and OoD performance computed on all 3000 architectures in our OoD-ViT-NAS benchmark. Measurements are computed on different OoD tasks including IN-C, IN-D, IN-P, IN-A, IN-R, IN-O, Stylized-IN, and IN-Sketch. }
	\label{fig:correlation-matrix-mean}       % Give a unique label
\end{figure}

Furthermore, we provide a comprehensive correlation between OoD accuracy and ID accuracy
in Fig.~\ref{fig:correlation-matrix-mean}.

\newpage

%% file: appx_sections/pareto.tex
\newpage

\section{Additional results on the analysis of OoD Performance of Pareto architectures for ID} \label{Sec:Appx_Pareto}

In the main paper, due to space constraints, we only provide the Pareto analysis results on a few representative OoD datasets. In this Appx. section, we provide additional results on this Pareto architecture analysis in Fig.~\ref{fig:pareto-small1}, ~\ref{fig:pareto-small2},~\ref{fig:pareto-base1}, ~\ref{fig:pareto-base2},~\ref{fig:pareto-tiny1},~\ref{fig:pareto-tiny2}. In the following scatter plots, blue dots \tikzcircle[bluecolor, fill=bluecolor]{2pt} represent architectures in the search space, while red dots \tikzcircle{2pt} represent the ID Pareto architectures. 

\newpage

%% file: appx_sections/proxy.tex
\section{Additional Results for Benchmarking Zero-cost Proxies} \label{Sec:Appx_Proxies}

In the main submission, we provides the comparison of Kendall $\tau$ ranking correlation between the ID/OoD accuracies and the zero-cost proxy values across all OoD datasets. In this section, we provides the correlation for each dataset for a detailed observation. The results can be found in Tab.~\ref{tab:proxy_IN_C},~\ref{tab:proxy_IN_P},~\ref{tab:proxy_IN_D},~\ref{tab:proxy_IN_Stylized},~\ref{tab:proxy_IN_Sketch},~\ref{tab:proxy_IN_R},~\ref{tab:proxy_IN_A},~\ref{tab:proxy_IN_O}. Our observations on individual OoD datasets are consistent with our findings in Sec.~\ref{Sec:Proxies}.

\begin{table}[h]
	\centering
	\caption{Comparison of Kendall $\tau$ ranking correlation between the ID/OoD accuracies and the zero-cost proxy values on ImageNet-C datasets in the Autoformer search space. \textbf{Bold} and \underline{underline} stands for the best and second, respectively.}
    \renewcommand{\arraystretch}{1.4}
	\resizebox{1.0\textwidth}{!}{%
		\begin{tabular}{@{}lcccc@{}}
			\toprule
            \multicolumn{1}{c}{\multirow{2}{*}{\textbf{Training-free NAS}}} & \multicolumn{2}{c}{\textbf{Originally Proposed For}}    & \multirow{2}{*}{\textbf{Correlation with ID Acc}} & \multirow{2}{*}{\textbf{Correlation with OoD Acc}} \\ \cline{2-3}
            \multicolumn{1}{c}{}                                   & \multicolumn{1}{c}{\textbf{Performance}} & \textbf{Architecture} &                                          &                                           \\ 
			\midrule
			Grasp \cite{wang2020picking}    & \multicolumn{1}{c}{ID Acc}          & CNNs           & 0.1490 $\pm$ 0.1951     & 0.1179 $\pm$ 0.1713 \\
   
			SNIP \cite{lee2018snip}      & \multicolumn{1}{c}{ID Acc}          & CNNs           & 0.3750 $\pm$ 0.3023     & 0.2371 $\pm$ 0.2579\\
   
			MeCo \cite{jiang2024meco}      & \multicolumn{1}{c}{ID Acc}          & CNNs           & 0.1440 $\pm$ 0.2371 & 0.0392 $\pm$ 0.1206\\

			CroZe \cite{ha2024generalizable}  & \multicolumn{1}{c}{Adv Robustness}  & CNNs           & 0.3823 $\pm$ 0.3046  & 0.2356 $\pm$ 0.2593 \\
   
			Jacobian \cite{hosseini2021dsrna} & \multicolumn{1}{c}{Adv Robustness}  & CNNs           & 0.1053 $\pm$ 0.1509 & 0.0894 $\pm$ 0.1369 \\

			DSS \cite{zhou2022training}      & \multicolumn{1}{c}{ID Acc}          & ViTs           & 0.4165 $\pm$ 0.3461 & 0.2454 $\pm$ 0.2630  \\
	
			AutoProx-A \cite{wei2024auto}& \multicolumn{1}{c}{ID Acc}         & ViTs           & 0.4023 $\pm$ 0.3827 & 0.2271 $\pm$ 0.2758  \\
   
            \rowcolor{lightgray}
			\#Param   & -               & -             & \underline{0.4607} $\pm$ \underline{0.3318} & \underline{0.2651} $\pm$ \underline{0.2546} \\
   
			\rowcolor{lightgray}
			\#Flops   & -               & -             & \textbf{0.4705} $\pm$ \textbf{0.3391} & \textbf{0.2656} $\pm$ \textbf{0.2572}  \\
   
			\bottomrule
		\end{tabular}}\label{tab:proxy_IN_C}
\end{table}

\begin{table}[h]
	\centering
	\caption{Comparison of Kendall $\tau$ ranking correlation between the ID/OoD accuracies and the zero-cost proxy values on ImageNet-P datasets in the Autoformer search space. \textbf{Bold} and \underline{underline} stands for the best and second, respectively.}
    \renewcommand{\arraystretch}{1.4}
	\resizebox{1.0\textwidth}{!}{%
		\begin{tabular}{@{}lcccc@{}}
			\toprule
            \multicolumn{1}{c}{\multirow{2}{*}{\textbf{Training-free NAS}}} & \multicolumn{2}{c}{\textbf{Originally Proposed For}}    & \multirow{2}{*}{\textbf{Correlation with ID Acc}} & \multirow{2}{*}{\textbf{Correlation with OoD Acc}} \\ \cline{2-3}
            \multicolumn{1}{c}{}                                   & \multicolumn{1}{c}{\textbf{Performance}} & \textbf{Architecture} &                                          &                                           \\ 
			\midrule
			Grasp \cite{wang2020picking}    & \multicolumn{1}{c}{ID Acc}          & CNNs           & 0.1490 $\pm$ 0.1951     & 0.1373 $\pm$ 0.1907 \\
   
			SNIP \cite{lee2018snip}      & \multicolumn{1}{c}{ID Acc}          & CNNs           & 0.3750 $\pm$ 0.3023     & 0.3652 $\pm$ 0.3306\\
   
			MeCo \cite{jiang2024meco}      & \multicolumn{1}{c}{ID Acc}          & CNNs           & 0.1440 $\pm$ 0.2371 & 0.1324 $\pm$ 0.2240\\

			CroZe \cite{ha2024generalizable}  & \multicolumn{1}{c}{Adv Robustness}  & CNNs           & 0.3823 $\pm$ 0.3046  & 0.3698 $\pm$ 0.3307 \\
   
			Jacobian \cite{hosseini2021dsrna} & \multicolumn{1}{c}{Adv Robustness}  & CNNs           & 0.1053 $\pm$ 0.1509 & 0.1142 $\pm$ 0.1590 \\

			DSS \cite{zhou2022training}      & \multicolumn{1}{c}{ID Acc}          & ViTs           & 0.4165 $\pm$ 0.3461 & 0.4128 $\pm$ 0.3621  \\
	
			AutoProx-A \cite{wei2024auto}& \multicolumn{1}{c}{ID Acc}         & ViTs           & 0.4023 $\pm$ 0.3827 & 0.3975 $\pm$ 0.3888  \\
   
            \rowcolor{lightgray}
			\#Param   & -               & -             & \underline{0.4607} $\pm$ \underline{0.3318} & \underline{0.4487} $\pm$ \underline{0.3475} \\
   
			\rowcolor{lightgray}
			\#Flops   & -               & -             & \textbf{0.4705} $\pm$ \textbf{0.3391} & \textbf{0.4592} $\pm$ \textbf{0.3547}  \\
   
			\bottomrule
		\end{tabular}}\label{tab:proxy_IN_P}
\end{table}

\begin{table}[h]
	\centering
	\caption{Comparison of Kendall $\tau$ ranking correlation between the ID/OoD accuracies and the zero-cost proxy values on ImageNet-D datasets in the Autoformer search space. \textbf{Bold} and \underline{underline} stands for the best and second, respectively.}
    \renewcommand{\arraystretch}{1.4}
	\resizebox{1.0\textwidth}{!}{%
		\begin{tabular}{@{}lcccc@{}}
			\toprule
            \multicolumn{1}{c}{\multirow{2}{*}{\textbf{Training-free NAS}}} & \multicolumn{2}{c}{\textbf{Originally Proposed For}}    & \multirow{2}{*}{\textbf{Correlation with ID Acc}} & \multirow{2}{*}{\textbf{Correlation with OoD Acc}} \\ \cline{2-3}
            \multicolumn{1}{c}{}                                   & \multicolumn{1}{c}{\textbf{Performance}} & \textbf{Architecture} &                                          &                                           \\ 
			\midrule
			Grasp \cite{wang2020picking}    & \multicolumn{1}{c}{ID Acc}          & CNNs           & 0.1490 $\pm$ 0.1951     & -0.0306 $\pm$ 0.0307 \\
   
			SNIP \cite{lee2018snip}      & \multicolumn{1}{c}{ID Acc}          & CNNs           & 0.3750 $\pm$ 0.3023     & \textbf{-0.0863} $\pm$ \textbf{0.0887}\\
   
			MeCo \cite{jiang2024meco}      & \multicolumn{1}{c}{ID Acc}          & CNNs           & 0.1440 $\pm$ 0.2371 & -0.0518 $\pm$ 0.0486\\

			CroZe \cite{ha2024generalizable}  & \multicolumn{1}{c}{Adv Robustness}  & CNNs           & 0.3823 $\pm$ 0.3046  & -0.0818 $\pm$ 0.0906 \\
   
			Jacobian \cite{hosseini2021dsrna} & \multicolumn{1}{c}{Adv Robustness}  & CNNs           & 0.1053 $\pm$ 0.1509 & -0.0090 $\pm$ 0.0318 \\

			DSS \cite{zhou2022training}      & \multicolumn{1}{c}{ID Acc}          & ViTs           & 0.4165 $\pm$ 0.3461 & -0.0743 $\pm$ 0.1131  \\
	
			AutoProx-A \cite{wei2024auto}& \multicolumn{1}{c}{ID Acc}         & ViTs           & 0.4023 $\pm$ 0.3827 & -0.0742 $\pm$ 0.1066  \\
   
            \rowcolor{lightgray}
			\#Param   & -               & -             & \underline{0.4607} $\pm$ \underline{0.3318} & \underline{-0.0837} $\pm$ \underline{0.1117} \\
   
			\rowcolor{lightgray}
			\#Flops   & -               & -             & 0.4705 $\pm$ 0.3391 & -0.0827 $\pm$ 0.1133  \\
   
			\bottomrule
		\end{tabular}}\label{tab:proxy_IN_D}
\end{table}

\begin{table}[h]
	\centering
	\caption{Comparison of Kendall $\tau$ ranking correlation between the ID/OoD accuracies and the zero-cost proxy values on Stylized-ImageNet datasets in the Autoformer search space. \textbf{Bold} and \underline{underline} stands for the best and second, respectively.}
    \renewcommand{\arraystretch}{1.4}
	\resizebox{1.0\textwidth}{!}{%
		\begin{tabular}{@{}lcccc@{}}
			\toprule
            \multicolumn{1}{c}{\multirow{2}{*}{\textbf{Training-free NAS}}} & \multicolumn{2}{c}{\textbf{Originally Proposed For}}    & \multirow{2}{*}{\textbf{Correlation with ID Acc}} & \multirow{2}{*}{\textbf{Correlation with OoD Acc}} \\ \cline{2-3}
            \multicolumn{1}{c}{}                                   & \multicolumn{1}{c}{\textbf{Performance}} & \textbf{Architecture} &                                          &                                           \\ 
			\midrule
			Grasp \cite{wang2020picking}    & \multicolumn{1}{c}{ID Acc}          & CNNs           & 0.1490 $\pm$ 0.1951     & 0.1413 $\pm$ 0.1854 \\
   
			SNIP \cite{lee2018snip}      & \multicolumn{1}{c}{ID Acc}          & CNNs           & 0.3750 $\pm$ 0.3023     & 0.3175 $\pm$ 0.2658\\
   
			MeCo \cite{jiang2024meco}      & \multicolumn{1}{c}{ID Acc}          & CNNs           & 0.1440 $\pm$ 0.2371 & 0.0937 $\pm$ 0.1739\\

			CroZe \cite{ha2024generalizable}  & \multicolumn{1}{c}{Adv Robustness}  & CNNs           & 0.3823 $\pm$ 0.3046  & 0.3091 $\pm$ 0.2620 \\
   
			Jacobian \cite{hosseini2021dsrna} & \multicolumn{1}{c}{Adv Robustness}  & CNNs           & 0.1053 $\pm$ 0.1509 & 0.1017 $\pm$ 0.1442 \\

			DSS \cite{zhou2022training}      & \multicolumn{1}{c}{ID Acc}          & ViTs           & 0.4165 $\pm$ 0.3461 & 0.3800 $\pm$ 0.3294  \\
	
			AutoProx-A \cite{wei2024auto}& \multicolumn{1}{c}{ID Acc}         & ViTs           & 0.4023 $\pm$ 0.3827 & 0.3619 $\pm$ 0.3573  \\
   
            \rowcolor{lightgray}
			\#Param   & -               & -             & \underline{0.4607} $\pm$ \underline{0.3318} & \underline{0.3899} $\pm$ \underline{0.3004} \\
   
			\rowcolor{lightgray}
			\#Flops   & -               & -             & \textbf{0.4705} $\pm$ \textbf{0.3391} & \textbf{0.3905} $\pm$ \textbf{0.3094}  \\
   
			\bottomrule
		\end{tabular}}\label{tab:proxy_IN_Stylized}
\end{table}

\begin{table}[h]
	\centering
	\caption{Comparison of Kendall $\tau$ ranking correlation between the ID/OoD accuracies and the zero-cost proxy values on ImageNet-Sketch datasets in the Autoformer search space. \textbf{Bold} and \underline{underline} stands for the best and second, respectively.}
    \renewcommand{\arraystretch}{1.4}
	\resizebox{1.0\textwidth}{!}{%
		\begin{tabular}{@{}lcccc@{}}
			\toprule
            \multicolumn{1}{c}{\multirow{2}{*}{\textbf{Training-free NAS}}} & \multicolumn{2}{c}{\textbf{Originally Proposed For}}    & \multirow{2}{*}{\textbf{Correlation with ID Acc}} & \multirow{2}{*}{\textbf{Correlation with OoD Acc}} \\ \cline{2-3}
            \multicolumn{1}{c}{}                                   & \multicolumn{1}{c}{\textbf{Performance}} & \textbf{Architecture} &                                          &                                           \\ 
			\midrule
			Grasp \cite{wang2020picking}    & \multicolumn{1}{c}{ID Acc}          & CNNs           & 0.1490 $\pm$ 0.1951     & 0.1395 $\pm$ 0.2299 \\
   
			SNIP \cite{lee2018snip}      & \multicolumn{1}{c}{ID Acc}          & CNNs           & 0.3750 $\pm$ 0.3023     & 0.3434 $\pm$ 0.3696\\
   
			MeCo \cite{jiang2024meco}      & \multicolumn{1}{c}{ID Acc}          & CNNs           & 0.1440 $\pm$ 0.2371 & 0.0896 $\pm$ 0.1112\\

			CroZe \cite{ha2024generalizable}  & \multicolumn{1}{c}{Adv Robustness}  & CNNs           & 0.3823 $\pm$ 0.3046  & 0.3610 $\pm$ 0.3616 \\
   
			Jacobian \cite{hosseini2021dsrna} & \multicolumn{1}{c}{Adv Robustness}  & CNNs           & 0.1053 $\pm$ 0.1509 & 0.0871 $\pm$ 0.1872 \\

			DSS \cite{zhou2022training}      & \multicolumn{1}{c}{ID Acc}          & ViTs           & 0.4165 $\pm$ 0.3461 & 0.3885 $\pm$ 0.4021  \\
	
			AutoProx-A \cite{wei2024auto}& \multicolumn{1}{c}{ID Acc}         & ViTs           & 0.4023 $\pm$ 0.3827 & 0.3599 $\pm$ 0.3939  \\
   
            \rowcolor{lightgray}
			\#Param   & -               & -             & \underline{0.4607} $\pm$ \underline{0.3318} & \textbf{0.4317} $\pm$ \textbf{0.4013} \\
   
			\rowcolor{lightgray}
			\#Flops   & -               & -             & \textbf{0.4705} $\pm$ \textbf{0.3391} & \underline{0.4060} $\pm$ \underline{0.3917}  \\
   
			\bottomrule
		\end{tabular}}\label{tab:proxy_IN_Sketch}
\end{table}

\begin{table}[h]
	\centering
	\caption{Comparison of Kendall $\tau$ ranking correlation between the ID/OoD accuracies and the zero-cost proxy values on ImageNet-R datasets in the Autoformer search space. \textbf{Bold} and \underline{underline} stands for the best and second, respectively.}
    \renewcommand{\arraystretch}{1.4}
	\resizebox{1.0\textwidth}{!}{%
		\begin{tabular}{@{}lcccc@{}}
			\toprule
            \multicolumn{1}{c}{\multirow{2}{*}{\textbf{Training-free NAS}}} & \multicolumn{2}{c}{\textbf{Originally Proposed For}}    & \multirow{2}{*}{\textbf{Correlation with ID Acc}} & \multirow{2}{*}{\textbf{Correlation with OoD Acc}} \\ \cline{2-3}
            \multicolumn{1}{c}{}                                   & \multicolumn{1}{c}{\textbf{Performance}} & \textbf{Architecture} &                                          &                                           \\ 
			\midrule
			Grasp \cite{wang2020picking}    & \multicolumn{1}{c}{ID Acc}          & CNNs           & 0.1490 $\pm$ 0.1951     & 0.1435 $\pm$ 0.1956 \\
   
			SNIP \cite{lee2018snip}      & \multicolumn{1}{c}{ID Acc}          & CNNs           & 0.3750 $\pm$ 0.3023     & 0.3576 $\pm$ 0.3416\\
   
			MeCo \cite{jiang2024meco}      & \multicolumn{1}{c}{ID Acc}          & CNNs           & 0.1440 $\pm$ 0.2371 & 0.1224 $\pm$ 0.1820\\

			CroZe \cite{ha2024generalizable}  & \multicolumn{1}{c}{Adv Robustness}  & CNNs           & 0.3823 $\pm$ 0.3046  & 0.3751 $\pm$ 0.3342 \\
   
			Jacobian \cite{hosseini2021dsrna} & \multicolumn{1}{c}{Adv Robustness}  & CNNs           & 0.1053 $\pm$ 0.1509 & 0.0935 $\pm$ 0.1518 \\

			DSS \cite{zhou2022training}      & \multicolumn{1}{c}{ID Acc}          & ViTs           & 0.4165 $\pm$ 0.3461 & 0.4364 $\pm$ 0.4047  \\
	
			AutoProx-A \cite{wei2024auto}& \multicolumn{1}{c}{ID Acc}         & ViTs           & 0.4023 $\pm$ 0.3827 & 0.4164 $\pm$ 0.4027  \\
   
            \rowcolor{lightgray}
			\#Param   & -               & -             & \underline{0.4607} $\pm$ \underline{0.3318} & \textbf{0.4773} $\pm$ \textbf{0.4179} \\
   
			\rowcolor{lightgray}
			\#Flops   & -               & -             & \textbf{0.4705} $\pm$ \textbf{0.3391} & \underline{0.4507} $\pm$ \underline{0.4028}  \\
   
			\bottomrule
		\end{tabular}}\label{tab:proxy_IN_R}
\end{table}

\begin{table}[h]
	\centering
	\caption{Comparison of Kendall $\tau$ ranking correlation between the ID/OoD accuracies and the zero-cost proxy values on ImageNet-A datasets in the Autoformer search space. \textbf{Bold} and \underline{underline} stands for the best and second, respectively.}
    \renewcommand{\arraystretch}{1.4}
	\resizebox{1.0\textwidth}{!}{%
		\begin{tabular}{@{}lcccc@{}}
			\toprule
            \multicolumn{1}{c}{\multirow{2}{*}{\textbf{Training-free NAS}}} & \multicolumn{2}{c}{\textbf{Originally Proposed For}}    & \multirow{2}{*}{\textbf{Correlation with ID Acc}} & \multirow{2}{*}{\textbf{Correlation with OoD Acc}} \\ \cline{2-3}
            \multicolumn{1}{c}{}                                   & \multicolumn{1}{c}{\textbf{Performance}} & \textbf{Architecture} &                                          &                                           \\ 
			\midrule
			Grasp \cite{wang2020picking}    & \multicolumn{1}{c}{ID Acc}          & CNNs           & 0.1490 $\pm$ 0.1951     & 0.1663 $\pm$ 0.1483 \\
   
			SNIP \cite{lee2018snip}      & \multicolumn{1}{c}{ID Acc}          & CNNs           & 0.3750 $\pm$ 0.3023     & 0.4588 $\pm$ 0.1477\\
   
			MeCo \cite{jiang2024meco}      & \multicolumn{1}{c}{ID Acc}          & CNNs           & 0.1440 $\pm$ 0.2371 & 0.2495 $\pm$ 0.0180\\

			CroZe \cite{ha2024generalizable}  & \multicolumn{1}{c}{Adv Robustness}  & CNNs           & 0.3823 $\pm$ 0.3046  & 0.4642 $\pm$ 0.1415 \\
   
			Jacobian \cite{hosseini2021dsrna} & \multicolumn{1}{c}{Adv Robustness}  & CNNs           & 0.1053 $\pm$ 0.1509 & 0.1050 $\pm$ 0.1172 \\

			DSS \cite{zhou2022training}      & \multicolumn{1}{c}{ID Acc}          & ViTs           & 0.4165 $\pm$ 0.3461 & 0.5930 $\pm$ 0.1021  \\
	
			AutoProx-A \cite{wei2024auto}& \multicolumn{1}{c}{ID Acc}         & ViTs           & 0.4023 $\pm$ 0.3827 & \textbf{0.6048} $\pm$ \textbf{0.0794}  \\
   
            \rowcolor{lightgray}
			\#Param   & -               & -             & \underline{0.4607} $\pm$ \underline{0.3318} & 0.5923 $\pm$ 0.1288 \\
   
			\rowcolor{lightgray}
			\#Flops   & -               & -             & \textbf{0.4705} $\pm$ \textbf{0.3391} & \underline{0.5959} $\pm$ \underline{0.1230}  \\
   
			\bottomrule
		\end{tabular}}\label{tab:proxy_IN_A}
\end{table}

\begin{table}[h]
	\centering
	\caption{Comparison of Kendall $\tau$ ranking correlation between the ID/OoD accuracies and the zero-cost proxy values on ImageNet-O datasets in the Autoformer search space. \textbf{Bold} and \underline{underline} stands for the best and second, respectively.}
    \renewcommand{\arraystretch}{1.4}
	\resizebox{1.0\textwidth}{!}{%
		\begin{tabular}{@{}lcccc@{}}
			\toprule
            \multicolumn{1}{c}{\multirow{2}{*}{\textbf{Training-free NAS}}} & \multicolumn{2}{c}{\textbf{Originally Proposed For}}    & \multirow{2}{*}{\textbf{Correlation with ID Acc}} & \multirow{2}{*}{\textbf{Correlation with OoD Acc}} \\ \cline{2-3}
            \multicolumn{1}{c}{}                                   & \multicolumn{1}{c}{\textbf{Performance}} & \textbf{Architecture} &                                          &                                           \\ 
			\midrule
			Grasp \cite{wang2020picking}    & \multicolumn{1}{c}{ID Acc}          & CNNs           & 0.1490 $\pm$ 0.1951     & 0.1503 $\pm$ 0.1638 \\
   
			SNIP \cite{lee2018snip}      & \multicolumn{1}{c}{ID Acc}          & CNNs           & 0.3750 $\pm$ 0.3023     & 0.3178 $\pm$ 0.3219\\
   
			MeCo \cite{jiang2024meco}      & \multicolumn{1}{c}{ID Acc}          & CNNs           & 0.1440 $\pm$ 0.2371 & 0.1049 $\pm$ 0.2860\\

			CroZe \cite{ha2024generalizable}  & \multicolumn{1}{c}{Adv Robustness}  & CNNs           & 0.3823 $\pm$ 0.3046  & 0.3277 $\pm$ 0.3307 \\
   
			Jacobian \cite{hosseini2021dsrna} & \multicolumn{1}{c}{Adv Robustness}  & CNNs           & 0.1053 $\pm$ 0.1509 & 0.0911 $\pm$ 0.1242 \\

			DSS \cite{zhou2022training}      & \multicolumn{1}{c}{ID Acc}          & ViTs           & 0.4165 $\pm$ 0.3461 & \underline{0.3546} $\pm$ \underline{0.3478}  \\
	
			AutoProx-A \cite{wei2024auto}& \multicolumn{1}{c}{ID Acc}         & ViTs           & 0.4023 $\pm$ 0.3827 & 0.3490 $\pm$ 0.3915  \\
   
            \rowcolor{lightgray}
			\#Param   & -               & -             & \underline{0.4607} $\pm$ \underline{0.3318} & \textbf{0.3583} $\pm$ \textbf{0.3559} \\
   
			\rowcolor{lightgray}
			\#Flops   & -               & -             & \textbf{0.4705} $\pm$ \textbf{0.3391} & 0.3445 $\pm$ 0.3887  \\
   
			\bottomrule
		\end{tabular}}\label{tab:proxy_IN_O}
\end{table}

%% file: appx_sections/vit_attribute_embedding.tex
\newpage

\section{Additional Figures of ViT Structural Attributes: Embedding Dimension} \label{Sec:Appx_ViTAttribute_Embedding}

\newpage

%% file: appx_sections/vit_attribute.tex
\section{Additional Analysis of ViT Structural Attributes on OoD Generalization} \label{Sec:Appx_ViTAttribute}

\subsection{Ablation Study on the Impact of ViT Architectural Attributes to OoD Generalization}
In this section, we demonstrate the effectiveness of each ViT architectural attribute on OoD accuracy from the ablation study perspective. All ablation studies are based on 1,000 ViT architectures sampled from Autoformer-Small search space in our OoD-ViT-NAS benchmark. Through our general analysis in Sec.~\ref{Sec:Robust_Arch} in the main and ablation study, we show that the embedding dimension has the highest impact among ViT architectural attributes, while network depth has a slight impact on OoD generalization.

\textbf{Experimental Setups.} We conduct the ablation study on the impact of ViT architectural attributes on  OoD generalization. Particularly, for each ablation study of one ViT architectural attribute, we vary that attribute while keeping all other attributes fixed. Then, we compute Kendall's $\tau$ rank correlation coefficient between each attribute and different OoD shifts. While we can directly adjust the depth and embedding dimension, adjusting MLP\_Ration and \#Head is challenging. This is because these two attributes for each ViT arch are in the form of a list with depth elements. Each element is selected among 3 choices. This results in a huge combination. To deal with this difficulty, we first compute the means of MLP\_Ration/\#Head for each architecture. Then, during the ablation study, we explore a range of values for MLP ratio and the number of heads (mean \#Head = $6 \pm 0.05$, mean MLP\_Ratio = $3.5 \pm 0.05$) to capture the impact of these more nuanced architectural variations. We note that this range of values is small, allowing us to approximately fix these two attributes.

\begin{figure}[H]
	\centering
	% Use the relevant command to insert your figure file.
	% For example, with the graphicx package use
	\includegraphics[keepaspectratio,scale=0.7]{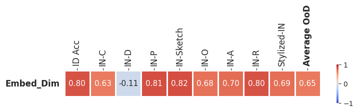}
	% figure caption is below the figure
	\caption{Kendall's $\tau$ rank correlation coefficient between varying Embed\_Dim and OoD accuracy.} %We fix Depth = $13$, mean \#Head = $6 \pm 0.05$, mean MLP\_Ratio = $3.5 \pm 0.05$. }
	
	\label{fig:ablation-edim}       % Give a unique label
\end{figure}

First, we assess the impact of embedding dimension on OoD generalization by fixing the configurations of all other ViT structural attributes (i.e., network depth = 13, mean \#Head = $6 \pm 0.05$, and mean MLP\_Ratio = $3.5 \pm 0.05$). The correlation results are shown in Fig.~\ref{fig:ablation-edim}. We observe an overall positive correlation of 0.65. This further supports our observation in Sec.~\ref{Sec:Robust_Arch} that increasing the embedding dimension generally could lead to better OoD performance across most OoD shifts for these ViT architectures.   %\textcolor{Red}{However, as shown in Figure~\ref{fig:embedding}, there are exceptions. For particularly severe OoD shifts, the correlation becomes close to zero or even negative, indicating that a larger embedding dimension might not always benefit OoD robustness.}\somayeh{We may include or not}

\begin{figure}[htb]
	\centering
	% Use the relevant command to insert your figure file.
	% For example, with the graphicx package use
	\includegraphics[keepaspectratio,scale=0.7]{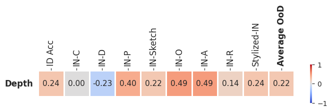}
	% figure caption is below the figure
	\caption{Kendall's $\tau$ rank correlation coefficient between varying network depth and all OoD accuracy.} %We fix Embed\_Dim = 384, mean \#Head = $6 \pm 0.05$, mean MLP\_Ratio = $3.5 \pm 0.05$.   }
	\label{fig:depth}       % Give a unique label
\end{figure}

To demonstrate how the number of layers (i.e., network depth) in a ViT architecture affects the model's OoD generalization, we fix other ViT architectural attributes (i.e., Embed\_Dim = $384$, mean \#Head = $6 \pm 0.05$, and mean MLP\_Ratio = $3.5 \pm 0.05$). 
% Fig.~\ref{fig:depth} depicts the ranking correlation for all architectures between depth and OoD accuracy. By analyzing the correlation between network depth and OoD performance in Fig.~ \ref{fig:depth}, 
The ranking correlation for all architectures between depth and OoD accuracy in Fig.~\ref{fig:depth} suggests a weak correlation between the ViT network depth and OoD accuracy. This observation is consistent with our finding in Sec.~\ref{Sec:Robust_Arch} that the depth has a minimal influence on OoD generalization.

\begin{figure}[htb]
	\centering
	% Use the relevant command to insert your figure file.
	% For example, with the graphicx package use
	\includegraphics[keepaspectratio,scale=0.7]{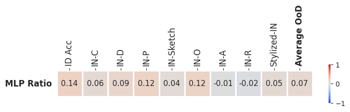}
	% figure caption is below the figure
	\caption{Kendall's $\tau$ rank correlation coefficient between mean MLP ratio and all OoD accuracy. We fix Embed\_Dim = $384$, Depth = $13$, and mean \#Head = $6 \pm 0.05$.   }
\label{fig:mlp-ablation}       % Give a unique label
\end{figure}

\begin{figure}[htb]
	\centering
	% Use the relevant command to insert your figure file.
	% For example, with the graphicx package use
	\includegraphics[keepaspectratio,scale = 0.7]{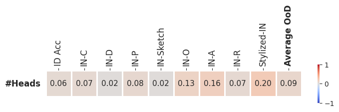}
	% figure caption is below the figure
	\caption{Kendall's $\tau$ rank correlation coefficient between the mean number of heads and all OoD accuracy. We fix Embed\_Dim = $384$, Depth = $13$, and mean MLP\_Ratio = $3.5 \pm 0.05$.   }
\label{fig:head-ablation}       % Give a unique label
\end{figure}

% These demonstrations align with our general observations in the main text, where we demonstrate an overall high impact of embedding dimension on OoD generalization, while network depth appears to have a slight influence on OoD generalization. 
For MLP\_Ratio and \#Heads, we observe that 
% our demonstration of 
the rank correlation coefficients of MLP\_Ratio and \#Heads in Fig.~\ref{fig:mlp-ablation} and \ref{fig:head-ablation}, respectively, reveals a weak correlation between the \#Heads/MLP\_Ratio and OoD generalization. This suggests that within the explored range, these two architectural attributes have a non-obvious impact on the model's OoD generalization. To delve deeper into this observation, Section~\ref{Sec:layerwise} presents a layer-wise analysis of these two architectural attributes in ViT models. %To explore this further, Section \ref{Sec:layerwise} we showcase a layer-wise analysis for these two architectural attributes of ViT models. 

\subsection{Layer-Wise Analysis}  \label{Sec:layerwise}

The number of heads and MLP ratio vary across layers, which allows for searching more diverse architectures. 
% The MLP ratio (or expansion ratio) expands the input embedding dimension by a factor at each layer.
In this section, we provide the layer-wise analysis of the influence of MLP\_Ratio/\#Heads in each layer on OoD generalization. 
% we design a layer-wise analysis by doing the ablation study on layer-level for MLP\_Ratio/\#Heads. 

\paragraph{Experimental Setups: } In Autoformer-Small search space, the MLP ratio at any given layer — referred to as the $i$-th layer — can be set to 3.0, 3.5, or 4. In addition to architectures in our benchmark, we create a set of 108 sampled architectures from Autoformer-Small. These architectures are fixed with three architectural design attributes: (Depth $= 12$, Embed\_Dim $= 320$, \#Heads for all layers $= 5$). 
With these parameters fixed, the total number of potential MLP\_Tatio configurations is still extensive (i.e., $3^{12}$), making it impractical to absolutely fix the MLP\_Ration in our ablation study.
To specifically assess the influence of the MLP\_Ration at the $i$-th layer, we further fix a constant MLP\_Ration across all other layers. For example, Fig.~\ref{fig:mlp-configuration} depicts the nine configurations of MLP\_Ration to analyze the impact of $5$-th layer.
% and then measured the OoD accuracy. 

We carry out a similar layer-wise analysis to demonstrate the effect of \#Heads at a specific layer ($i$-th layer). In addition to architectures in our benchmark, we create a set of 108 sampled from Autoformer-Small. These architectures are fixed with three architectural design attributes: (Depth = $12$, Embed\_Dim =$320$, \#MLP\_Ratio for all layers =$3.0$). To specifically assess the influence of the \#Head at the $i$-th layer, we further fix a constant \#Head across all other layers. For example, Fig.~\ref{fig:mlp-configuration} depicts the nine configurations of \#Head to analyze the impact of $3$-th layer.
% , where the 108 sampled architectures are fixed with three architectural design attributes, as follows:

\begin{figure}[!htb]
\centering
% Use the relevant command to insert your figure file.
% For example, with the graphicx package use
\includegraphics[keepaspectratio,width=1.\textwidth]{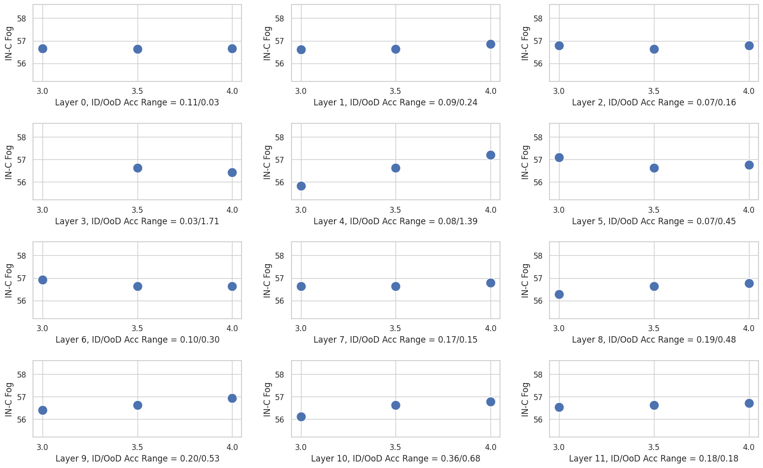}
% figure caption is below the figure
\caption{	A visualization on the effect of changing MLP ratio per layer to OoD accuracy in 108 architectures sampled from Autoformer-Small in the layer-wise study. To evaluate the effect of layer $i$-th, we fix the MLP ratio $=4.0$ for the remaining layers. We observe that a slightly higher OoD accuracy range can be obtained by changing the MLP ratio at layers 4 and 5. 
}% we show that higher ID/OoD accuracy can be obtained by increasing the embedding dimension of ViT blocks. Our observation shows that increasing the network depth does not have a high impact on OoD robustness.   }
\label{fig:mlp-configuration1}       % Give a unique label
\end{figure}

\textbf{Results.} The layer-wise analysis for MLP\_Ratio are shown in Fig.~\ref{fig:mlp-configuration1}. Each sub-figure demonstrates the change in OoD Accuracy when varying MLP\_Ratio at a particular layer. While increasing it improved OoD accuracy in some layers, it decreased it in others. This aligns with our observations in Sec.~\ref{Sec:Robust_Arch}, suggesting no clear overall impact of MLP\_Ratio on OoD generalization.

\begin{figure}[H]
\centering
% Use the relevant command to insert your figure file.
% For example, with the graphicx package use
\includegraphics[keepaspectratio,width=1.\textwidth]{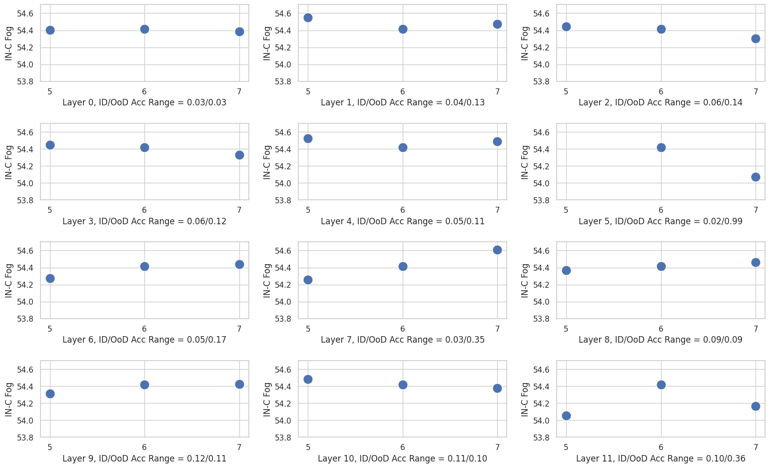}
% figure caption is below the figure
\caption{A visualization on the effect of changing \#Heads per layer to OoD accuracy in 108 architectures sampled from Autoformer-Small in the layer-wise study. To evaluate the effect of layer $i$-th, we fix the \#Heads $=6$ for the remaining layers.  }% we show that higher ID/OoD accuracy can be obtained by increasing the embedding dimension of ViT blocks. Our observation shows that increasing the network depth does not have a high impact on OoD robustness.   }
\label{fig:head-configuration}       % Give a unique label
\end{figure}

The layer-wise analysis for \#Head is illustrated in Fig.~\ref{fig:head-configuration}. Similar to our observation in the layer-wise analysis for MLP\_Ratio, the impact of \#Head is non-obvious, which is consistent with our finding in Sec.~\ref{Sec:Robust_Arch}.
% The results are shown in Tab.~\ref{tab:ablation-mlp}, where we demonstrate that changing the MLP ratio at layers \#4 and \#5 has a slightly increased range of mean OoD robust accuracy. The detailed visualization of three different layer-wise fixed MLP\_Ratio $=\{3.0, 3.5, 4.0\}$ is represented in Fig.~\ref{fig:mlp-configuration1},\ref{fig:mlp-configuration2}, and\ref{fig:mlp-configuration3}. 

% \input{ablation-mlp.tex} 

% \input{tables/ablation-head.tex} 

% Overall, our layer-wise demonstration reveals no obvious trend in the correlation between \#Heads and OoD accuracy. The average Kendall rank correlation coefficients for each OoD shift and the average OoD accuracy fluctuate across layers without a consistent pattern. This suggests that the number of heads at each layer does not have a straightforward or predictable impact on OoD generalization. Consequently, other factors, such as embedding dimension, may play a more significant role in influencing OoD performance. 

%% file: appx_sections/reproducibility.tex
\section{Reproducibility} \label{Sec:Appx_Reproducibility}

\begin{figure}[H]
	\centering
	% Use the relevant command to insert your figure file.
	% For example, with the graphicx package use
	\includegraphics[scale=0.5]{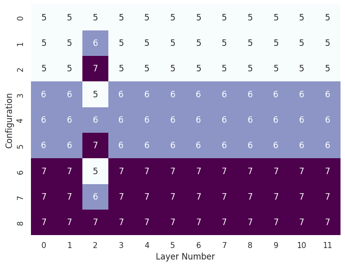}
	% figure caption is below the figure
	\caption{The possible configurations of \#Heads at $3$rd layer of 108 architectures sampled from \textit{Autoformer-Small} in layer-wise analysis. }
\label{fig:num-head-configuration}       % Give a unique label
\end{figure}

\begin{figure}[H]
	\centering
	% Use the relevant command to insert your figure file.
	% For example, with the graphicx package use
	\includegraphics[scale=0.5]{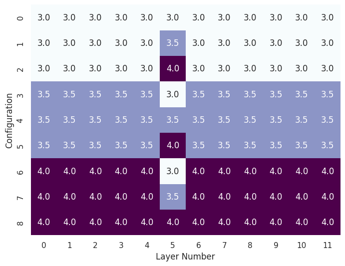}
	% figure caption is below the figure
	\caption{The possible configurations of MLP ratio at $5$-th layer of 108 architectures sampled from Autoformer-Small in the layer-wise study. }% we show that higher ID/OoD accuracy can be obtained by increasing the embedding dimension of ViT blocks. Our observation shows that increasing the network depth do s not have a high impact on OoD robustness.   }
\label{fig:mlp-configuration}       % Give a unique label
\end{figure}

\subsection{Hyper-parameters}
\begin{table}[H]
    \centering
    \caption{Hyper-parameters for Evaluation on $8$ common large-scale OoD datasets. In total, we evaluate 3,000 diverse ViT architectures in our OoD-NAS-ViT benchmark sampled from Autoformer-Tiny/Small/Base search spaces \cite{chen2021autoformer}. Input resolution is set to $224\times224$ pixels, with mean and standard deviation normalization applied using ImageNet statistics ($mean=[0.485, 0.456, 0.406], std=[0.229, 0.224, 0.225]$). Transformations follow the Standard ImageNet preprocessing, including resize and center crop.}
    \renewcommand{\arraystretch}{2.4}
	\resizebox{1.0\textwidth}{!}{%
    \begin{tabular}{rcccccccc}
    \toprule
                      & \textbf{IN-C} & \textbf{IN-P} & \textbf{IN-A} & \textbf{IN-O} & \textbf{IN-R} & \textbf{IN-Sketch} & \textbf{Stylized-IN} & \textbf{IN-D} \\
    \bottomrule
\textbf{Batch Size}        & 256        & 256        & 256        & 64         & 256        & 256             & 256               & 100        \\
\textbf{Number of workers} & 10         & 10         & 10         & 4          & 10         & 10              & 10                & 10         \\
    \bottomrule
    \end{tabular}}\label{tab:parameter}
\end{table}

To ensure reproducibility, we provide a detailed description of the hyper-parameters used for evaluating 3,000 ViT architectures in our OoD-NAS-ViT benchmarks
% Vision Transformer architectures sampled from AutoFormer-Tiny, AutoFormer-Small and AutoFormer-Base Search Space \cite{chen2021autoformer} 
on $8$ common large-scale OoD datasets: ImageNet-C \cite{hendrycks2019benchmarking}, ImageNet-A \cite{hendrycks2021natural} \cite{hendrycks2019benchmarking}, ImageNet-O \cite{hendrycks2021natural}, ImageNet-P \cite{hendrycks2019benchmarking}, ImageNet-D \cite{zhang2024imagenet}, ImageNet-R \cite{hendrycks2021many}, ImageNet-Sketch \cite{wang2019learning}, and Stylized ImageNet \cite{geirhos2018imagenet}. The evaluation for ImageNet-D, ImageNet-O and Stylized ImageNet strictly follows previous works \cite{zhang2024imagenet,hendrycks2021natural,geirhos2018imagenet} to ensure consistency and comparability. The details of the evaluation 
% on OoD datasets procedure 
are shown in Table \ref{tab:parameter}. The evaluated architectures are sampled on AutoFormer-Tiny/Small/Base Search Spaces \cite{chen2021autoformer}.
%, and we measured both ID accuracy and out-of-distribution performances.

\subsection{Compute Resource}
All our experiments are conducted using NVIDIA RTX A6000 GPUs.
% , each equipped with 40GB of memory
% To optimize parallel processing capabilities, 
We utilized $2$ GPUs for each experiment. 
% Our primary objective was to evaluate Vision Transformer architectures sampled from the AutoFormer-Tiny, AutoFormer-Small, and AutoFormer-Base search spaces \cite{chen2021autoformer} on $8$ common large-scale OoD datasets. 
The substantial computational resources required for these evaluations underscore the complexity and scale of our work. Detailed information on the GPU-hour consumed for all experiments to construct our OoD-ViT-NAS Benchmark can be found in Table \ref{tab:compute_resource}. In total, the experiments demanded a significant investment of approximately 3900 GPU-hours, reflecting the extensive computational effort involved.
\begin{table}[H]
    \centering
    \caption{GPU-Hour for Computational Resources. In total, we evaluate 3,000 diverse ViT architectures in our OoD-NAS-ViT benchmark sampled from Autoformer-Tiny/Small/Base search spaces \cite{chen2021autoformer}.}
    \renewcommand{\arraystretch}{2.4}
	\resizebox{1.0\textwidth}{!}{
    \begin{tabular}{rccccccccc}
    \toprule
                  & \textbf{IN-C} & \textbf{IN-P} & \textbf{IN-A} & \textbf{IN-O} & \textbf{IN-R} & \textbf{IN-Sketch} & \textbf{Stylized-IN} & \textbf{IN-D} & \textbf{Total} \\
    \bottomrule
\textbf{GPU-Hour} & 2958                & 672                 & 12                  & 93                  & 28                  & 53                       & 37                         & 50                  & \textbf{3903} \\
    \bottomrule
    \end{tabular}} \label{tab:compute_resource}
\end{table}

\newpage
\begin{figure}[H]
	\centering
	% Use the relevant command to insert your figure file.
	% For example, with the graphicx package use
	\includegraphics[width=1.\textwidth]{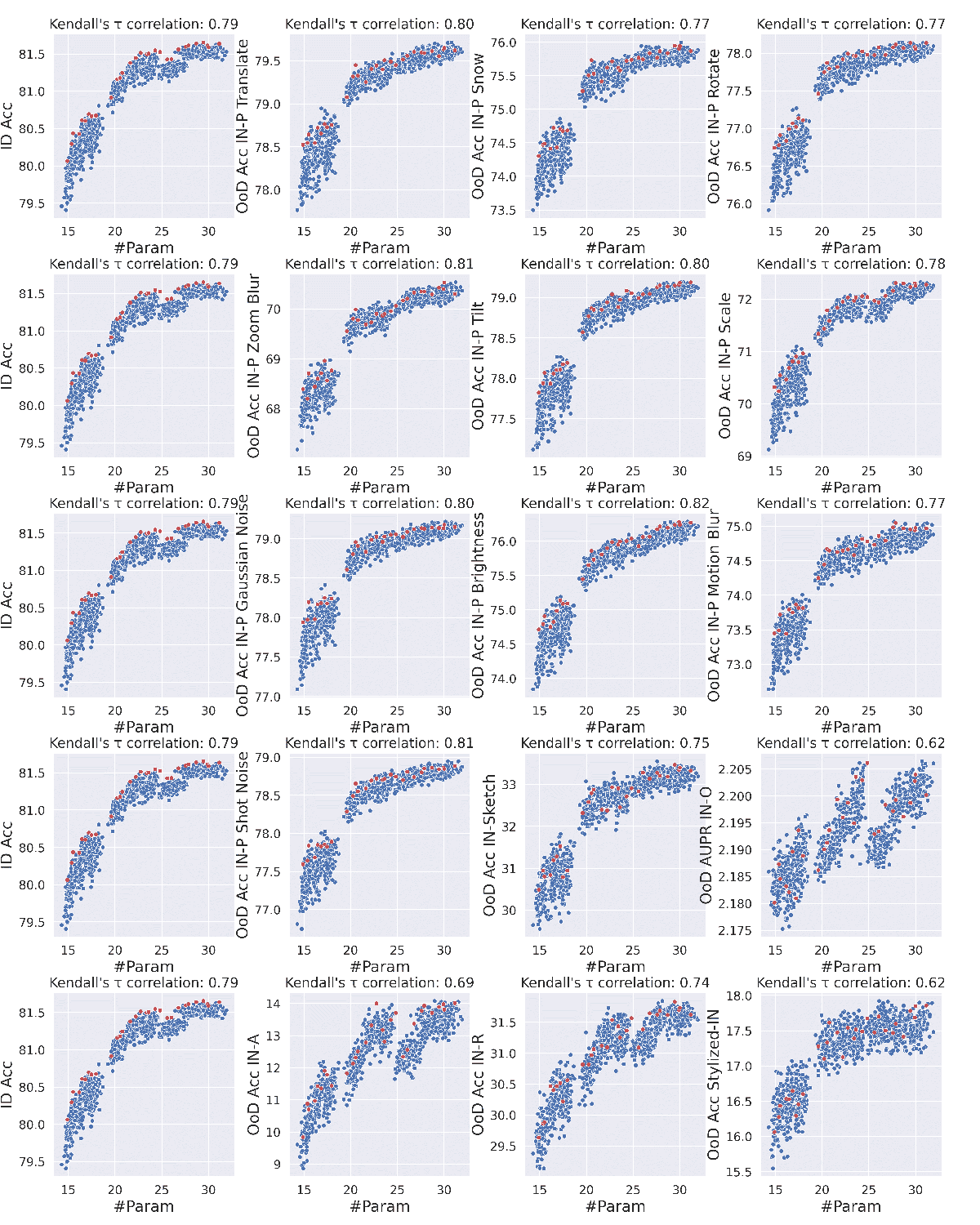}
	% figure caption is below the figure
	\caption{As in Figure \ref{fig:pareto}, we show that lower OoD accuracy can be obtained for the higher ID accuracy in the Pareto architectures of \textit{Autoformer-Small}. The left panels show the ID accuracy, and each panel on columns 2 to 4 shows results from the OoD accuracy of IN-P, IN-A, IN-R, IN-Sketch, Stylized-IN, and AUPR of IN-O.   }
	\label{fig:pareto-small2}       % Give a unique label
\end{figure}

\begin{figure}[H]
	\centering
	% Use the relevant command to insert your figure file.
	% For example, with the graphicx package use
	\includegraphics[width=1.\textwidth]{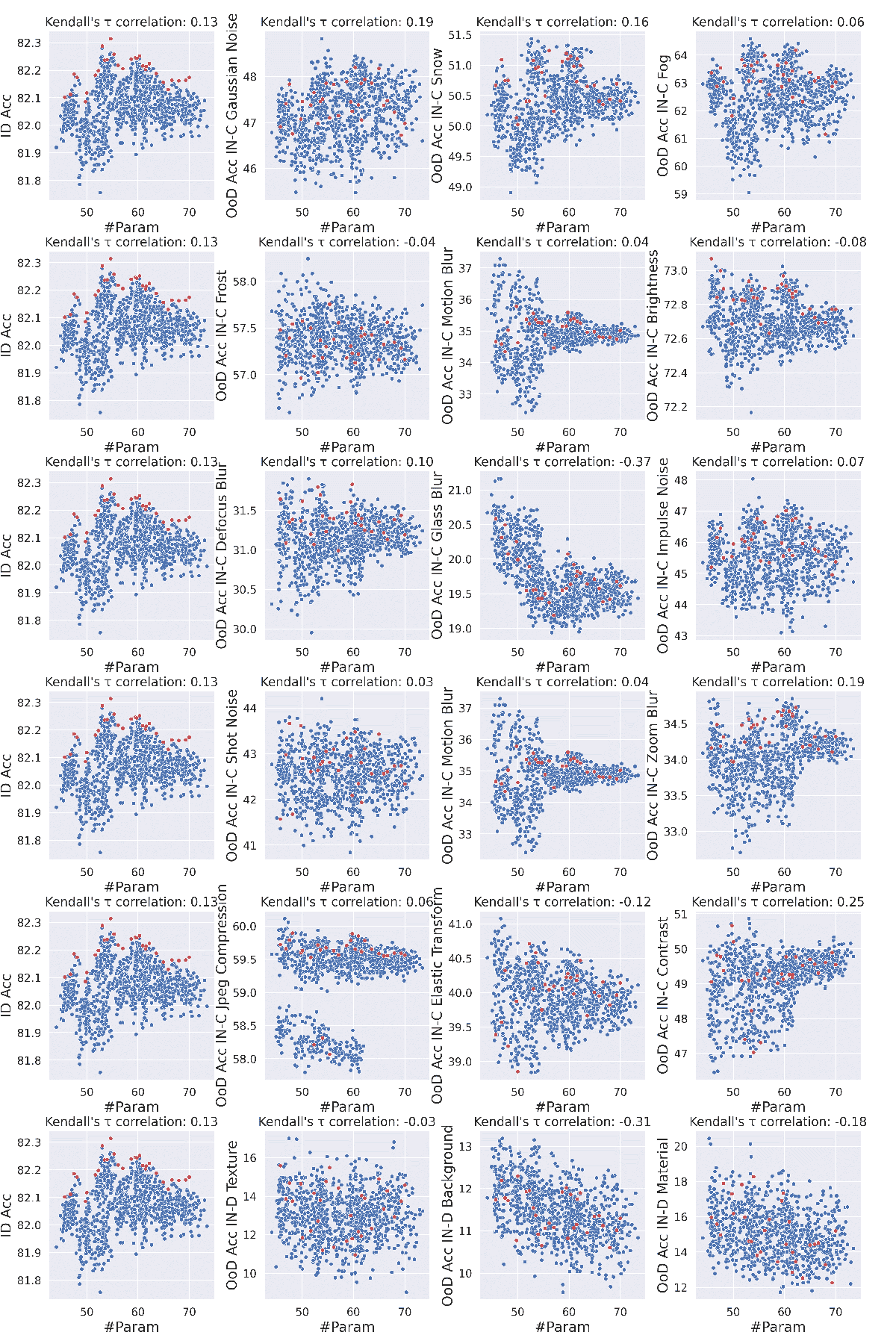}
	% figure caption is below the figure
	\caption{Visualization of Pareto architectures in \textit{Autoformer-Base}. The left panels show the ID accuracy, and each panel on columns 2 to 4 shows results from the OoD accuracy of IN-C common corruptions and IN-D.   }
	\label{fig:pareto-base1}       % Give a unique label
\end{figure}

\begin{figure}[H]
	\centering
	% Use the relevant command to insert your figure file.
	% For example, with the graphicx package use
	\includegraphics[width=1.\textwidth]{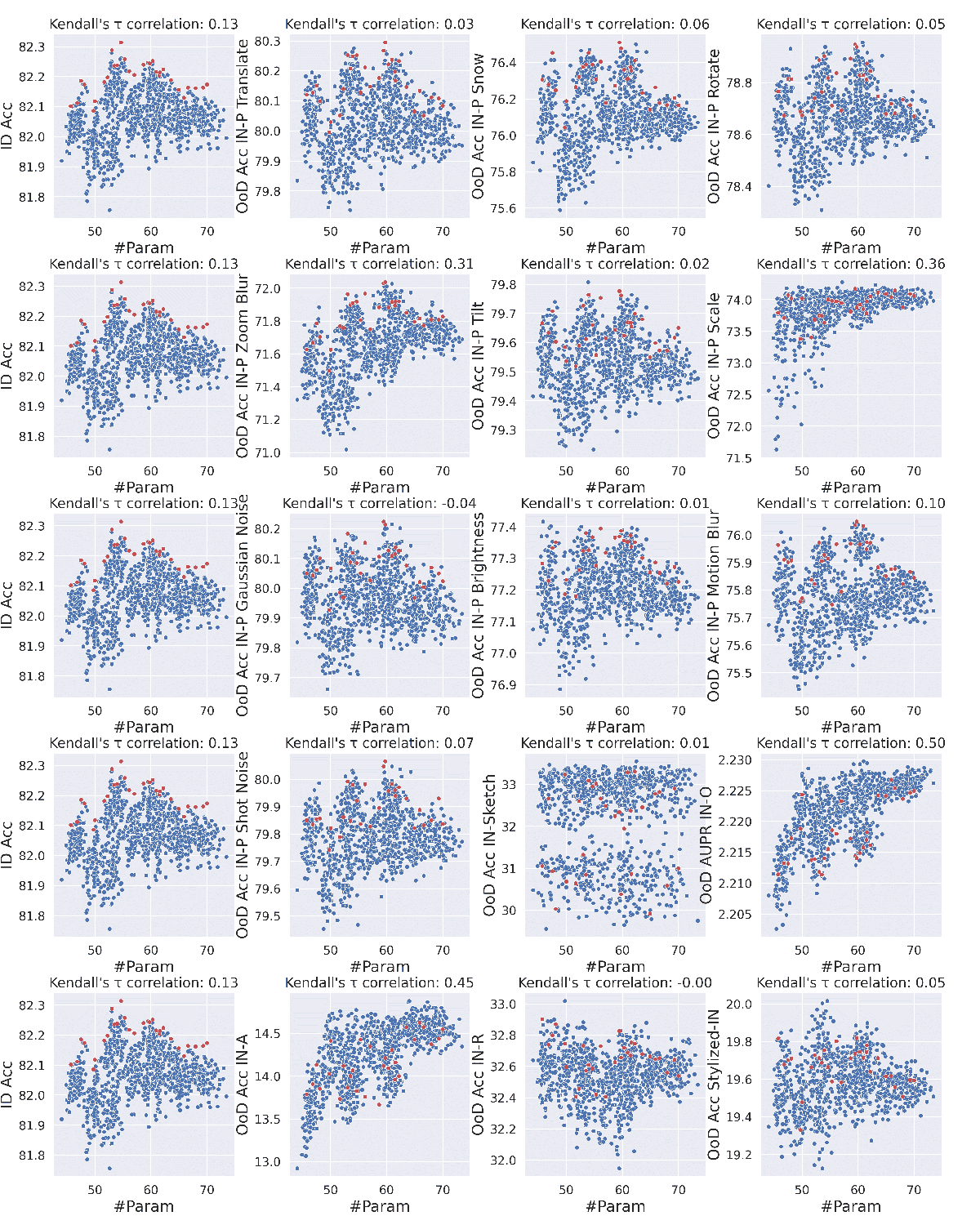}
	% figure caption is below the figure
	\caption{Visualization of Pareto architectures in \textit{Autoformer-Base}. The left panels show the ID accuracy, and each panel on columns 2 to 4 shows results from the OoD accuracy of IN-P, IN-A, IN-R, IN-Sketch, Stylized-IN, and AUPR of IN-O.   }
	\label{fig:pareto-base2}       % Give a unique label
\end{figure}

\begin{figure}[H]
	\centering
	% Use the relevant command to insert your figure file.
	% For example, with the graphicx package use
	\includegraphics[width=1.\textwidth]{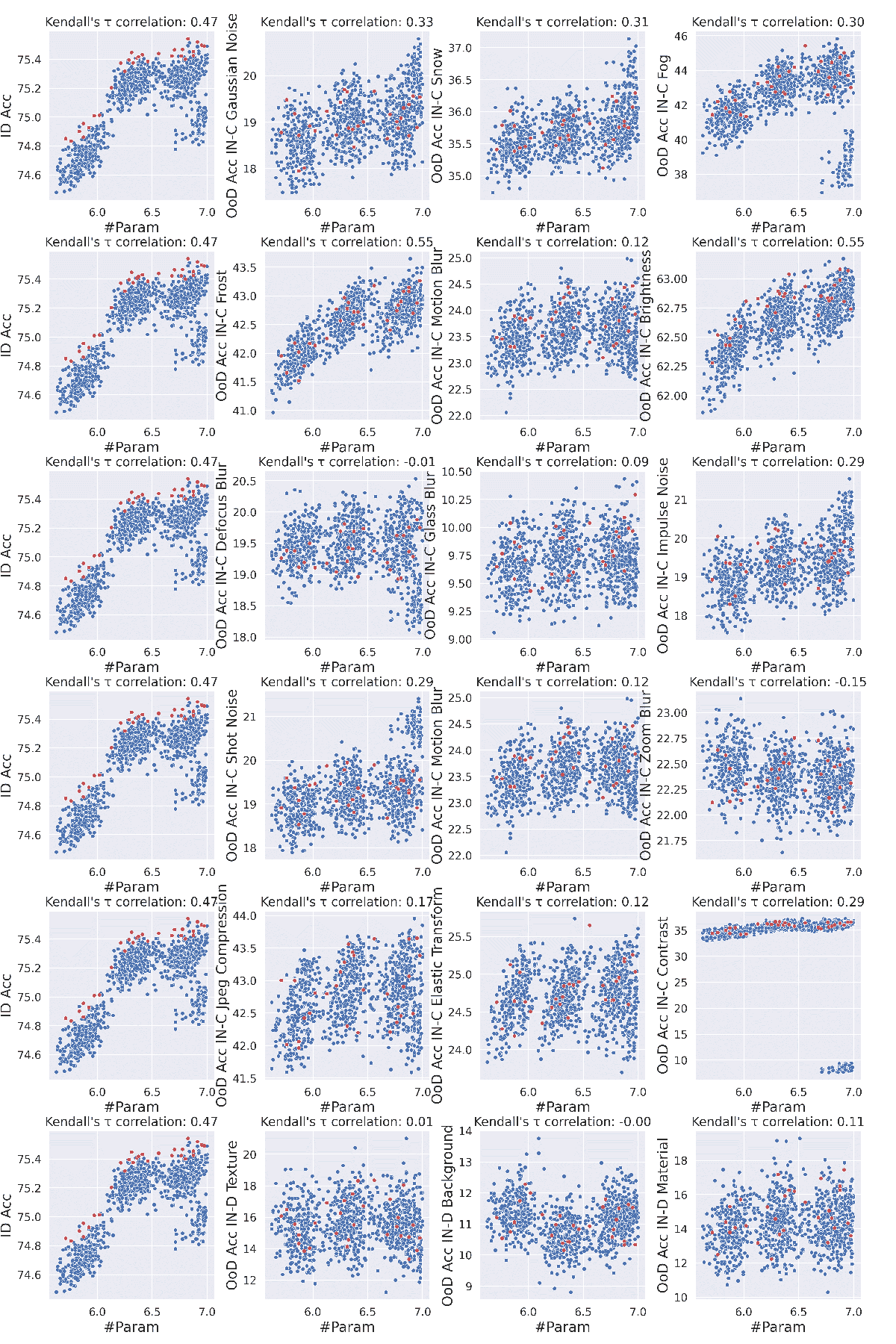}
	% figure caption is below the figure
	\caption{Visualization of Pareto architectures in \textit{Autoformer-Tiny}. The left panels show the ID accuracy, and each panel on columns 2 to 4 shows results from the OoD accuracy of IN-C common corruptions and IN-D.   }
	\label{fig:pareto-tiny1}       % Give a unique label
\end{figure}

\begin{figure}[H]
	\centering
	% Use the relevant command to insert your figure file.
	% For example, with the graphicx package use
	\includegraphics[width=1.\textwidth]{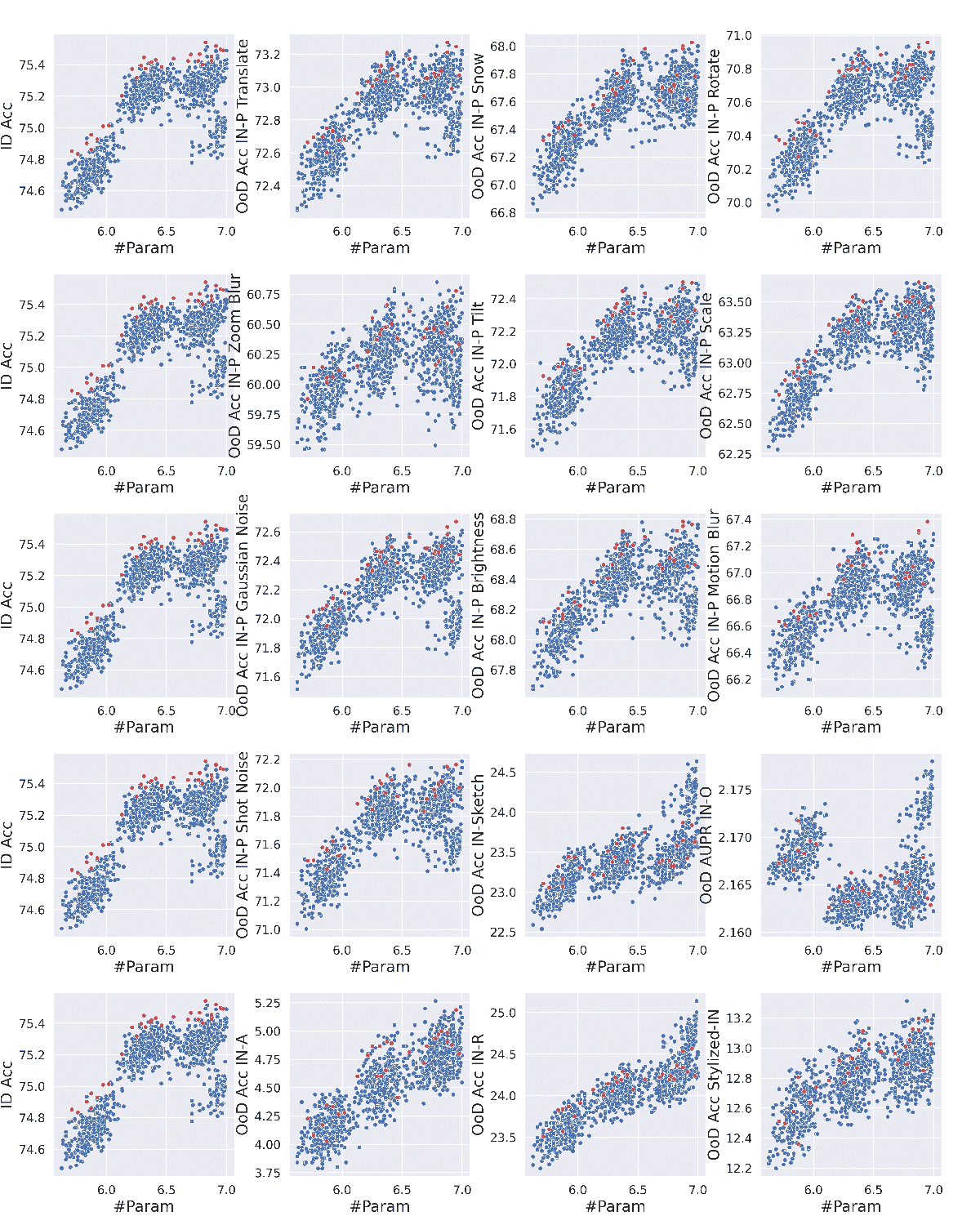}
	% figure caption is below the figure
	\caption{Visualization of Pareto architectures in \textit{Autoformer-Tiny}. The left panels show the ID accuracy, and each panel on columns 2 to 4 shows results from the OoD accuracy of IN-P, IN-A, IN-R, IN-Sketch, Stylized-IN, and AUPR of IN-O.   }
	\label{fig:pareto-tiny2}       % Give a unique label
\end{figure}

\begin{figure}[H]
	\centering
	% Use the relevant command to insert your figure file.
	% For example, with the graphicx package use
\includegraphics[width=1.\textwidth]{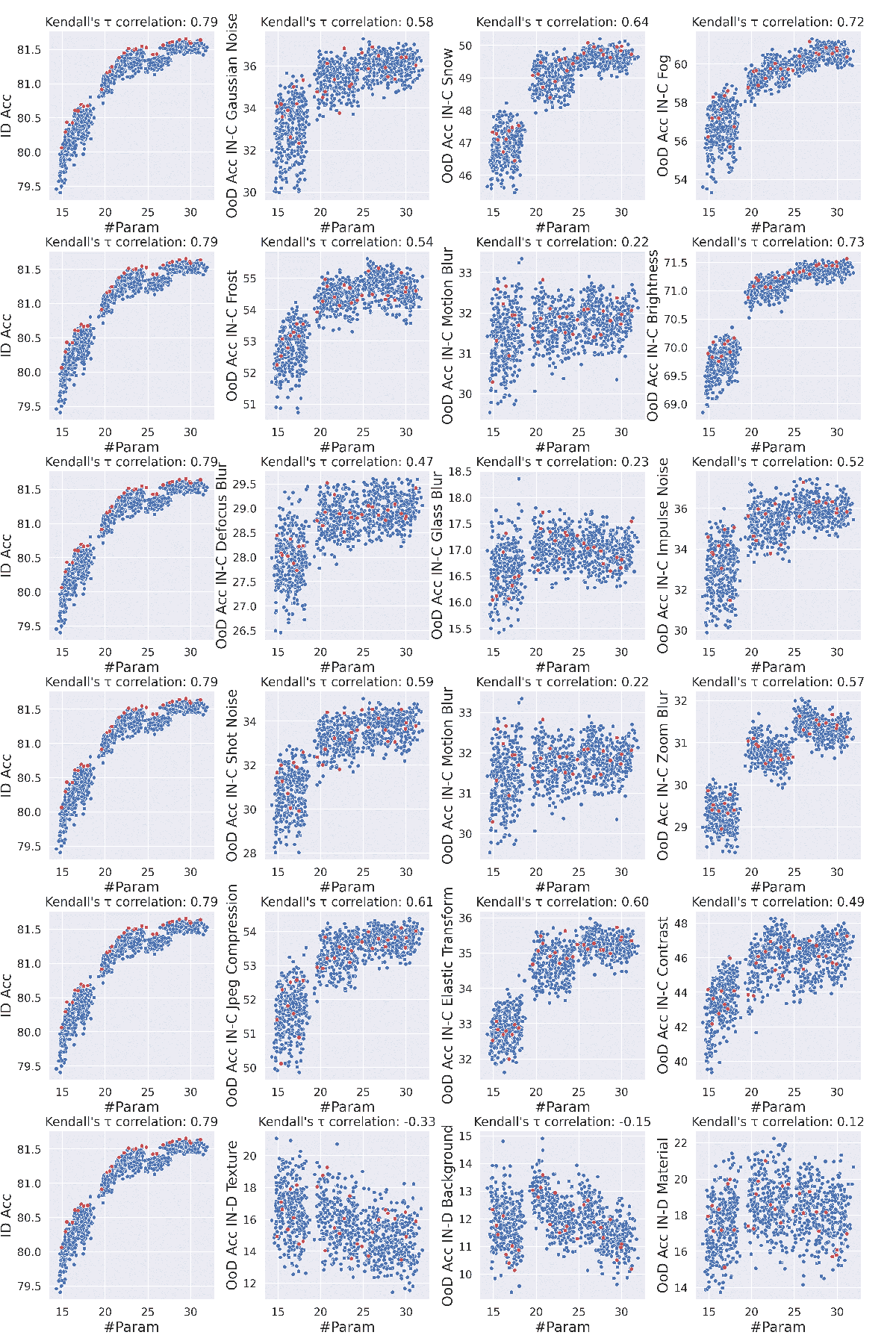}
	% figure caption is below the figure
	\caption{As in Figure \ref{fig:pareto}, we show that lower OoD accuracy can be obtained for the higher ID accuracy in the Pareto architectures of \textit{Autoformer-Small}. The left panels show the ID accuracy, and each panel in columns 2 to 4 shows results from OoD accuracy of IN-C common corruptions and IN-D. We observe that architectural designs have a greater effect on OoD accuracy than ID accuracy, especially when OoD shifts become more severe.  }
	\label{fig:pareto-small1}       % Give a unique label
\end{figure}

\begin{figure}[H]
	\centering
	% Use the relevant command to insert your figure file.
	% For example, with the graphicx package use
	\includegraphics[width=1.\textwidth]{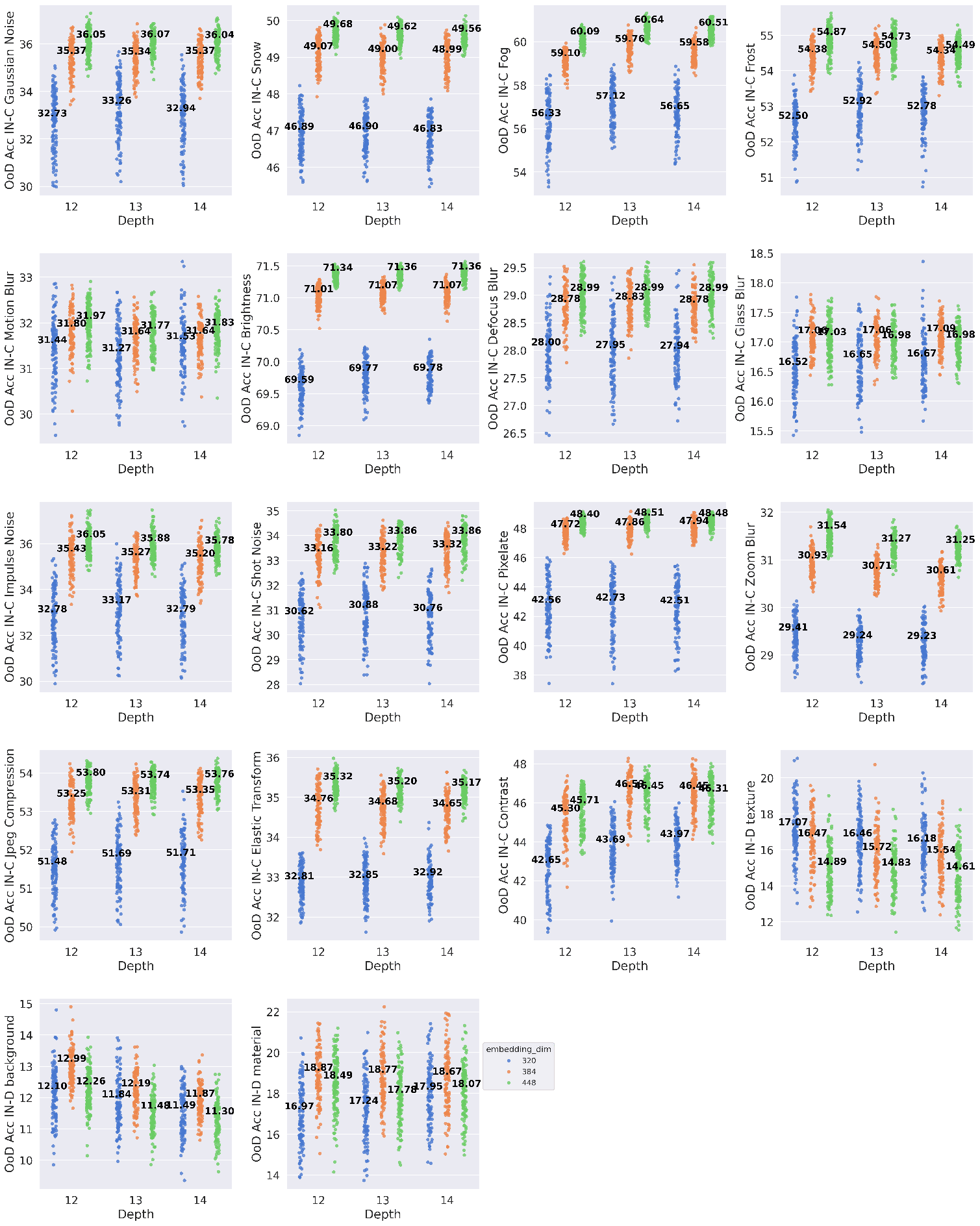}
	% figure caption is below the figure
	\caption{As in Figure \ref{fig:embedding}, we show the potential impact of embedding dimension (Embed\_Dim) on OoD generalization in ViTs architectures sampled from Autoformer-Small. The numbers denote the average OoD performance, and The data points with blue \tikzcircle[color320, fill=color320]{2pt}, orange \tikzcircle[color384, fill=color384]{2pt}, and green \tikzcircle[color448, fill=color448]{2pt} colours represent ViT architectures with the embedding dimension of 320, 384, and 448, respectively. Each panel shows results from the OoD accuracy of IN-C common corruptions and IN-D. }
	\label{fig:embedding-all-small1}       % Give a unique label
\end{figure}

\begin{figure}[H]
	\centering
	% Use the relevant command to insert your figure file.
	% For example, with the graphicx package use
	\includegraphics[width=1.\textwidth]{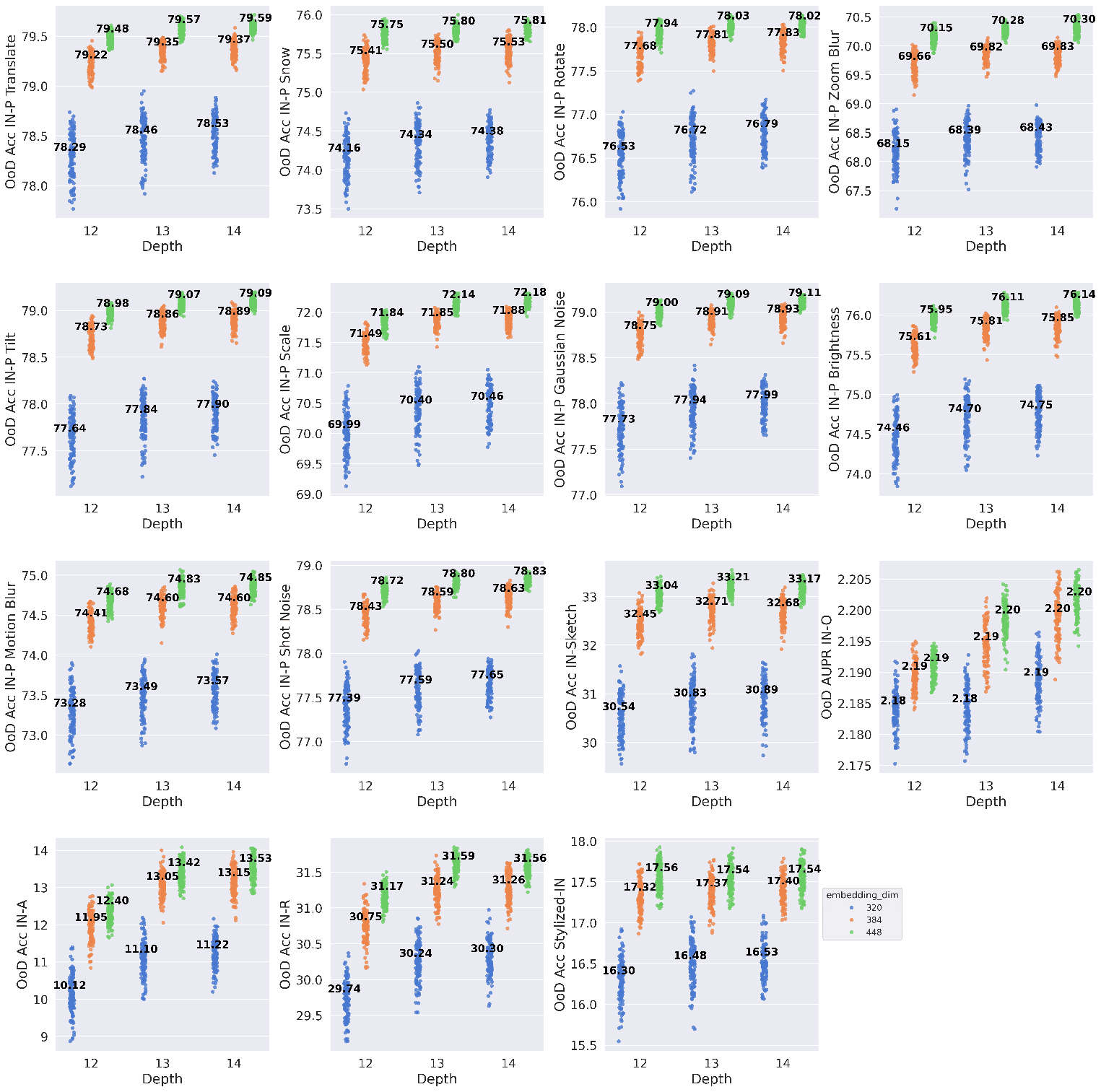}
	% figure caption is below the figure
	\caption{As in Figure \ref{fig:embedding}, we show the potential impact of embedding dimension (Embed\_Dim) on OoD generalization in ViTs architectures sampled from Autoformer-Small. The numbers denote the average OoD performance, and The data points with blue \tikzcircle[color320, fill=color320]{2pt}, orange \tikzcircle[color384, fill=color384]{2pt}, and green \tikzcircle[color448, fill=color448]{2pt} colours represent ViT architectures with the embedding dimension of 320, 384, and 448, respectively. Each panel shows results from the OoD accuracy of IN-P, IN-A, IN-R, IN-Sketch, Stylized-IN, and AUPR of IN-O. }
	\label{fig:embedding-all-small2}       % Give a unique label
\end{figure}